\documentclass[11pt]{article}
\pdfoutput=1

\usepackage{graphicx, verbatim}
\usepackage{amssymb}
\usepackage{alltt}
\usepackage{amsmath, amssymb, amsfonts, amscd, xspace, pifont, amsthm}
\usepackage{mathrsfs}
\newtheorem{theorem}{Theorem}
\newtheorem{lemma}[theorem]{Lemma}
\newtheorem{proposition}{Proposition}

\newtheorem{definition}{Definition}

\usepackage[round]{natbib}

\def\threeImages#1#2#3#4#5#6#7#8#9 
{
\centerline{\hfill\makebox[#2]{#3}\hfill\makebox[#5]{#6}\hfill\makebox[#8]{#9}\hfill}
\centerline{\hfill
\includegraphics[width=#2]{#1}
\hfill
\includegraphics[width=#5]{#4}
\hfill
\includegraphics[width=#8]{#7}
\hfill}
}

\def\twoImages#1#2#3#4#5#6 
{
\centerline{\hfill\makebox[#2]{#3}\hfill\makebox[#5]{#6}\hfill}
\centerline{\hfill
\includegraphics[width=#2]{#1}
\hfill
\includegraphics[width=#5]{#4}
\hfill}
}
\theoremstyle{remark}
\newtheorem{remark}{Remark}

\DeclareMathOperator*{\argmax}{arg\,max}

\usepackage{multirow}

\usepackage{comment}
\usepackage[ruled,vlined]{algorithm2e}

\begin{document}
\title{Variational Estimators of the Degree-corrected
	Latent Block Model for Bipartite Networks}
\author{
	Yunpeng Zhao, Ning Hao, and Ji Zhu
  } 

\maketitle

\begin{abstract}
Bipartite graphs are ubiquitous across various scientific and engineering fields. Simultaneously grouping the two types of nodes in a bipartite graph via biclustering represents a fundamental challenge in network analysis for such graphs. The latent block model (LBM) is a commonly used model-based tool for biclustering. However, the effectiveness of the LBM is often limited by the influence of row and column sums in the data matrix.
To address this limitation, we introduce the degree-corrected latent block model (DC-LBM), which accounts for the varying degrees in row and column clusters, significantly enhancing performance on real-world data sets and simulated data. We develop an efficient variational expectation-maximization algorithm by creating closed-form solutions for parameter estimates in the M steps. 
Furthermore, we prove the label consistency and the rate of convergence of the variational estimator under the DC-LBM, allowing the expected graph density to approach zero as long as the average expected degrees of rows and columns approach infinity when the size of the graph increases.	
	
	\end{abstract}

{\it Keywords:}  biclustering, bipartite graph, identifiability, label consistency,  variational expectation-maximization

\section{Introduction} \label{sec:intro}

Biclustering or coclustering, first considered by \cite{hartigan1972direct}, is an unsupervised learning task that simultaneously clusters  the rows and columns of a rectangular data matrix. Biclustering is a machine learning technique with many applications, such as in genomics \citep{cheng2000biclustering,pontes2015biclustering}, recommender systems  \citep{alqadah2015biclustering}, and text mining \citep{de2007applying,orzechowski2016text}. Similar to standard cluster analysis, an exhaustive search  for all possible partitions of rows and columns is intractable due to the exponential growth in the number of possible partitions with the increase in row and column numbers. Many popular biclustering methods employ greedy algorithms to find the local optimal partition according to certain criteria. Examples include Minimum Sum-Squared Residue Coclustering (MSSRCC) \citep{cho2004minimum} and Large Average Submatrices (LAS) \citep{shabalin2009finding}. For a systematic review  and comparison of typical biclustering algorithms, readers are referred to \cite{padilha2017systematic}.

Mixture models, such as Gaussian mixture models \citep{Fraley2002} for continuous data and Bernoulli mixture models for binary data \citep{celeux1991clustering}, provide a natural probabilistic framework for standard cluster analysis, where each observation is associated with a latent cluster label that can be inferred by estimating posterior probabilities given the data. Similarly, model-based approaches have also been developed for biclustering. 
We focus on one of the most popular models for biclustering — the latent block model (LBM), first proposed by \cite{govaert2003clustering}. The LBM is a natural generalization of mixture models to the ``two-dimensional'' case, where the probability distribution of each entry of the data matrix depends on both the row and column labels.

The expectation-maximization (EM) algorithm is the most widely-used algorithm for fitting a mixture model. However, the E step of the EM algorithm for the LBM becomes intractable due to the complex dependence structure among entries of the data matrix and cluster labels \citep{govaert2008block}. To overcome this computational obstacle, \cite{govaert2003clustering} proposed the block classification EM (CEM) algorithm. It includes an additional C step that assigns hard cluster labels based on the estimated posterior probabilities in the current iteration. Once the hard row labels are obtained, the column labels can be updated using a standard EM algorithm, and vice versa.
\cite{govaert2006fuzzy} introduced the fuzzy block criterion, which avoids the conversion of posterior probabilities into hard labels. The criterion function was later reinterpreted within the framework of variational EM algorithms and named block EM \citep{govaert2008block}, a technique we will adapt in this paper.  In this approach, the block EM algorithm maximizes a lower bound of the log-likelihood function, referred to as the variational approximation. It imposes a constraint that allows for factoring the posterior distribution of row and column labels.

We study the biclustering problem for data matrices in which an entry represents the relationship between the corresponding row and column objects. In other words, we consider biclustering on bipartite (two-mode) networks with single or multiple edges. This formulation is closely related to another widely-studied area---community detection for one-mode networks. Although developed almost independently from biclustering, the models and algorithms for community detection, and challenges that they face are similar to biclustering. The stochastic block model (SBM), first proposed by \cite{Holland83}, is the best studied model in the community detection literature. The SBM can be viewed as the analogue  of the LBM for symmetric binary networks although there is little overlap between literatures of the two models until recently.  \cite{Bickel&Chen2009} established the first theoretical framework to study the consistency of estimated labels under the SBM, and in particular they proved the consistency of profile likelihood estimators. The theoretical framework was extended by \cite{flynn2020profile}  to biclustering for a wide range of
data modalities, including binary, count, and continuous observations.
When fitting the SBM, the E step of the classical EM algorithm is intractable as in the LBM. 
Various computationally efficient approaches have been proposed for fitting the SBM, including, but not limited to, variational approximation \citep{daudin2008mixture,bickel2013asymptotic}, pseudo likelihood \citep{amini2013pseudo}, split likelihood \citep{wang2021efficient}, and profile-pseudo likelihood \citep{wang2021fast}. Readers are referred to \cite{abbe2017community} and \cite{zhao2017survey} for surveys on the computational and theoretical advances for the SBM and related models.

The LBM has a notable limitation in practical applications to bipartite networks: it tends to cluster rows with similar row degrees (i.e. row sums) and columns with similar column degrees (i.e. column sums) together. This issue is also observed in the SBM used for symmetric networks. \cite{Karrer10} proposed the degree-corrected stochastic block model (DC-SBM) that includes an additional set of parameters controlling  expected degrees. The degree parameters were usually estimated implicitly in the community detection literature, which is partially due to the model identifiability issue. For example, \cite{amini2013pseudo} proposed the conditional pseudo likelihood which models the number of edges in a block as a multinomial variable conditional on the  observed degrees. A similar approach was used   in the split likelihood method \citep{wang2021efficient}.

In this paper, we propose a degree-corrected latent block model (DC-LBM) to accommodate degree heterogeneity in biclustering. Instead of using any surrogate, we take a direct approach: we adapt the block EM algorithm to estimate all parameters in the original form of the DC-LBM, including degree parameters, without ad-hoc modification. 
We show that the observed row and column degrees, up to constants, are exactly the maximizers for the corresponding degree parameters in the M step given any probability assignment on the cluster labels, if we model the entries of the data matrix as independent Poisson variables conditional on the cluster labels.
The estimates of the degree parameters therefore remain constant in the algorithm, which results an elegant and efficient estimating procedure under the variational EM framework.   

The theoretical contribution of the present paper is the establishment of the label consistency and the rate of convergence of the variational estimator under the DC-LBM. \cite{brault2020consistency} proved the consistency of the maximum likelihood estimator and the variational estimator under the classical LBM by  showing both the marginal likelihood and the
variational approximation are asymptotically equivalent to the complete data likelihood. We take a new approach by directly proving  the variational approximation uniformly converges to its population version and the true cluster labels are a well-separated maximizer of the population version, which implies label consistency. A key ingredient in the proof is
a uniform concentration inequality over probability assignments of cluster labels.
The proof can accommodate degree parameters and requires weaker conditions. In particular, we allow that the expected graph density goes to zero provided that both the average expected row and column degrees go to infinity as the size of the network increases, which is a typical condition for label consistency for symmetric networks \citep{Bickel&Chen2009,zhao2012consistency}.

This paper focuses on likelihood-based approaches to biclustering on bipartite networks. Spectral clustering, co-clustering, and their numerous variants, as computationally efficient non-likelihood-based approaches, have been widely applied to network data. This class of methods typically involves the construction of various types of graph Laplacians, followed by the application of eigendecomposition or singular value decomposition to the constructed matrices. Spectral clustering methods were first proposed for undirected networks \citep{Rohe2011,  jin2015fast, Sarkar_role, lei2015consistency}, and soon be generalized to directed networks \citep{rohe2016co,Ji2020,Zhang2022_directed_embedding}. It is particularly worth mentioning that biclustering is closely related to co-clustering in directed networks when the row and column labels are presumed to be different; that is, they represent distinct communities for ``sending'' versus ``receiving'',  as in the setup of \cite{rohe2016co}. The only difference compared to the bipartite graph problem is the assumption of square matrices for directed graphs.

In terms of related theoretical results,
\cite{zhao2012consistency} extended the theoretical framework of \cite{Bickel&Chen2009}  and proved a general theorem for label consistency under the DC-SBM, where the degree parameters can take a finite number of possible values. \cite{amini2013pseudo,wang2021efficient,wang2021fast} also considered the DC-SBM but the theoretical analyses focused on the classical SBM. \cite{flynn2020profile} extended the theoretical framework of profile likelihood methods \citep{Bickel&Chen2009} to biclustering. \cite{mariadassou2015convergence} proposed a unified framework for
studying the convergence of the posterior distribution of cluster labels  under both the SBM and LBM, assuming known parameter values.

The rest of this paper is organized as follows. In Section \ref{sec:model}, we introduce the DC-LBM with the Poisson distribution. In Section \ref{sec:alg}, we propose a variational EM algorithm for DC-LBM, with a key property stating that the row and column degrees maximize the objective function in the M step, given any probability assignment on the cluster labels. Section \ref{sec:theory} addresses  asymptotic properties. We establish the consistency of label estimation and the rate of convergence under the DC-LBM with both the Poisson and the Bernoulli distributions on edges. In Section \ref{sec:simu}, we compare the performance of the proposed method with other popular biclustering algorithms in various setups. In Section \ref{sec:data}, we apply the proposed method to a benchmark data set for biclustering---the MovieLens data set. All technical proofs are provided in the appendix. Additionally, we include the analysis of an SMS spam data set in the appendix.

\section{Model}\label{sec:model}
We introduce the degree-corrected latent block model (DC-LBM) for bipartite networks in this section. Consider an adjacency matrix $A=[A_{ij}]$ with $m$ rows and $n$ columns, where each $A_{ij}$ is a non-negative integer that represents  multiple edges  or an edge with an integer weight from $i$ to $j$.

We assume that the row (resp. column) indices are partitioned into $K$ (resp. $L$) latent clusters. Denote the cluster labels on rows by ${z}=(z_1,...,z_m)^{T}$ and the labels on columns by ${w}=(w_1,...,w_n)^{T}$. Assume  $z_1,...,z_m$ are independently and identically distributed (i.i.d.) multinomial variables with $\textnormal{Multi}(1,{\pi}=(\pi_1,...,\pi_K)^{T})$. Similarly, $w_1,...,w_n$ are i.i.d. multinomial variables with $\textnormal{Multi}(1,{\rho}=(\rho_1,...,\rho_L)^{T})$.
In addition to the cluster structure, we use ``degree parameters'' ${\theta}=(\theta_1,...,\theta_m)^{T}$ and ${\lambda}=(\lambda_1,...,\lambda_n)^{T}$ to model the propensities of row and column objects to form links. Both cluster labels ${z},{w}$ and degree parameters ${\theta},{\lambda}$ are unknown, and throughout the paper we treat ${z},{w}$ as latent variables and ${\theta},{\lambda}$ as parameters.

Conditional on the cluster labels, $\{ A_{ij} \}$ are independent Poisson variables with mean $ \{\theta_i \lambda_j\mu_{z_i w_j}\}$ where $\mu=[\mu_{kl}]$ is a $K$-by-$L$ matrix. Hence, the joint likelihood of ${z},{w}$ and $A$ is
\begin{align*}
P({z},{w},A ;{\pi},{\rho},{\theta},{\lambda},  \mu) =\left(\prod_{i=1}^m \pi_{z_i}\right)\left( \prod_{j=1}^n  \rho_{w_j}\right) \prod_{i=1}^m \prod_{j=1}^n e^{-\theta_i \lambda_j \mu_{z_i w_j}}  \frac{(\theta_i \lambda_j \mu_{z_i w_j})^{A_{ij}}}{A_{ij}!} .
\end{align*}
Certainly, the parameters $\theta$, $\lambda$ and $\mu$ must meet specific constraints for identifiability. We will discuss the resolution to this issue in Sections \ref{sec:alg} and \ref{sec:theory}.

We assume the Poisson distribution in the model  for the convenience of algorithm development. The  DC-SBM for symmetric networks, first proposed by \cite{Karrer10}, also assumes the Poisson distribution. As will be shown in simulation studies and data analysis, the model performs well for networks with binary edges.  We prove label consistency and the rate of convergence under both the Bernoulli and Poisson models in Section \ref{sec:theory}.   Furthermore,  the model reduces to the classical LBM when $\theta_i\equiv 1,\lambda_j\equiv 1, i=1,...,m,j=1,...,n$.

Since the cluster labels are latent, we consider the marginal likelihood of $A$
\begin{align}
P(A; {\pi},{\rho},{\theta},{\lambda},  \mu)= \sum_{{z}\in \Omega_{{z}}} \sum_{{w}\in \Omega_{{w}}} P({z},{w},A ;{\pi},{\rho},{\theta},{\lambda},  \mu), \label{marginal}
\end{align}
where $\Omega_{{z}}= \{1,...,K\}^m$ and $\Omega_{{w}}= \{1,...,L\}^n$. Note that a brute-force calculation of the summation in \eqref{marginal} is intractable because the number of terms grows exponentially with $m$ and $n$, and $ P({z},{w},A ;{\pi},{\rho},{\theta},{\lambda},  \mu)$ cannot factor under these sums.  The classical EM algorithm is intractable for the same reason. Specifically, the E step involves a sum of $K^m L^n$ terms as in \eqref{marginal}, owing to the dependence structure among the variables $A_{ij}$. 

\section{Variational expectation-maximization algorithm}\label{sec:alg}
To address the computational challenges, we employ a strategy akin to the variational EM algorithm used for the classical LBM, as outlined by \cite{govaert2006fuzzy,govaert2008block}. This algorithm reframes the EM algorithm as a coordinate ascent method, where the E-step is treated as maximization across a range of probability measures. When the E-step is computationally daunting, a solution can be sought by maximizing over a \textit{constrained} space instead.

\subsection{General framework}
We begin with the general framework of the algorithm. Due to Jensen's inequality,\footnote{We use the convention $0 \log 0=0$, which is consistent with $\lim_{x \rightarrow 0} x \log x=0$.}
\begin{align*}
\log P(A)= & \log \sum_{{z}\in \Omega_{{z}}} \sum_{{w}\in \Omega_{{w}}} P({z},{w},A; \Phi) \geq   \sum_{{z}\in \Omega_{{z}}} \sum_{{w}\in \Omega_{{w}}} q({z},{w}) \log \left \{  \frac{P({z},{w},A; \Phi)}{ q({z},{w})} \right \},
\end{align*}
where $\Phi=({\pi},{\rho},{\theta},{\lambda},\mu)$, and $q$ is a probability measure over $\Omega_{{z}}\times \Omega_{{w}}$ satisfying $$\sum_{{z}\in \Omega_{{z}}} \sum_{{w}\in \Omega_{{w}}}  q({z},{w})=1.$$ Equality holds if $ q({z},{w})$ is equal to the posterior probability $\mathbb{P}({z},{w}|A)$. The standard EM algorithm is to iteratively update $\Phi$ and $q$.

As mentioned in Section \ref{sec:model}, a brute-force calculation of $\sum_{{z}\in \Omega_{{z}}} \sum_{{w}\in \Omega_{{w}}}$ is intractable. To resolve this issue,  \cite{govaert2008block} proposed to impose the constraint $q({z},{w})=q_1({z}) q_2({w})$ with  $\sum_{{z}\in \Omega_{{z}}} q_1({z})=1$ and $\sum_{{w}\in \Omega_{{w}}} q_2({w})=1$. Let
\begin{align}
J(q_1,q_2,\Phi) = \sum_{{z}\in \Omega_{{z}}} \sum_{{w}\in \Omega_{{w}}} q_1({z}) q_2({w}) \log \left \{  \frac{P({z},{w},A; \Phi)}{q_1({z}) q_2({w})} \right \}. \label{main_obj}
\end{align}
Conditional on $A$, the latent variables ${z}$ and  ${w}$ are not  independent. Therefore, $q_1({z}) q_2({w})$ is not exactly the posterior probability $\mathbb{P}({z},{w}|A)$. But their dependence is weak for large $m$ and $n$. The intuition behind the constraint $q({z},{w})=q_1({z}) q_2({w})$ is that if the detection of clusters is consistent (as we will prove in Section \ref{sec:theory}),  the posterior distribution of $({z},{w})$ will eventually concentrate on a single realization---the true cluster labels. In other words, the posterior distribution of $({z},{w})$ can be  approximated by a Dirac measure, allowing for factorization.


The variational EM algorithm iteratively updates the parameters $\Phi$ (M step) and the probability measures $q_1$ and $q_2$ (E step). We specify the two steps in the following subsections.

\subsection{M step}\label{sec:M}

The degree parameters in the likelihood were typically indirectly addressed in the literature. Instead of explicitly including the degree parameters, likelihood functions conditional on observed degrees were used, such as conditional pseudo-likelihood \citep{amini2013pseudo} and conditional split likelihood \citep{wang2021efficient}. This approach is based on the observation that the conditional distribution of independent Poisson variables, given their sum, follows a multinomial distribution \citep{amini2013pseudo}. 

In this paper, we take a different and more direct approach. Instead of using any surrogate \citep{amini2013pseudo,wang2021efficient}, in the M step we rigorously maximize $J(q_1,q_2,\Phi)$ over all parameters including ${\theta}$ and ${\lambda}$ given $q_1$ and $q_2$. A key observation here is that the row and column degrees are global optimizers for  ${\theta}$ and ${\lambda}$ for any given $q_1$ and $q_2$ (Proposition \ref{thm:stationary}).

Let $$\mathbb{P}_{q_1}(z_i=k)=\sum_{{z}\in \Omega_{{z}}} q_1(z_1,...,z_{i-1},k,z_{i+1},...,z_m), \,\, i=1,...,m,k=1,...,K,$$ and $$\mathbb{P}_{q_2}(w_j=l)=\sum_{{w}\in \Omega_{{w}}} q_2(w_1,...,w_{j-1},l,w_{j+1},...,w_n), \,\, j=1,...,n,l=1,...,L.$$ 
Let $d_i^r=\sum_{j=1}^n A_{ij}\,\, (i=1,...,m)$ and $d_j^c=\sum_{i=1}^m A_{ij}\,\, (j=1,...,n)$ be the row and column degrees, respectively.

We have the following result:
\begin{proposition}\label{thm:stationary}
	For any fixed $q_1$ and $q_2$, $\hat\Phi=(\hat{{\pi}},\hat{{\rho}},\hat{{\theta}},\hat{{\lambda}},\hat{\mu}) $ defined below is a global maximizer of $J(q_1,q_2,\Phi)$.
	\begin{align*}
	& \hat{\theta}_i=d_i^r, \,\, i=1,...,m, \\
    &\hat{\lambda}_j=d_j^c, \,\, j=1,...,n, \\
	&	 \hat{\mu}_{kl}= \frac{\sum_{i=1}^m \sum_{j=1}^n  A_{ij}  \mathbb{P}_{q_1}(z_i=k) \mathbb{P}_{q_2}(w_j=l)}{\sum_{i=1}^m \sum_{j=1}^n d_i^r d_j^c \mathbb{P}_{q_1}(z_i=k) \mathbb{P}_{q_2}(w_j=l) }, \,\, k=1,...,K, l=1,...,L, \\
	&    	\hat{\pi}_k =  \frac{1}{m} \sum_{i=1}^m  \mathbb{P}_{q_1}(z_i=k), \,\, k=1,...,K, \\
    & \hat{\rho}_l =   \frac{1}{n} \sum_{j=1}^n  \mathbb{P}_{q_2}(w_j=l), \,\, l=1,...,L.
	\end{align*}
	Moreover, if $\mathbb{P}_{q_1}(z_i=k)\ne0$ for all $i$ and $k$, and $\mathbb{P}_{q_2}(w_j=l)\ne0$ for all $j$ and $l$, all maximizers are of the form $(\hat{{\pi}},\hat{{\rho}},e^{c_1}\hat{{\theta}},e^{c_2}\hat{{\lambda}},e^{-c_1-c_2}\hat\mu)$, where $c_1$ and $c_2$ are two constants.
\end{proposition}

Proposition \ref{thm:stationary} indicates that for generic $q_1$ and $q_2$, the maximizer is unique up to two multiplicative scalars $e^{c_1}$ and $e^{c_2}$. (Please refer to our proof on the uniqueness result for all possible $q_1$ and $q_2$.)
For the convenience of the theoretical study in Section \ref{sec:theory}, we will choose
\begin{align*}
& \hat{\theta}_i  = \frac{d_i^r}{n\sqrt{D}}, \,\, i=1,...,m, \,\, \hat{\lambda}_j  = \frac{d_j^c}{m\sqrt{D}}, \,\, j=1,...,n, \,\, \textnormal{where }  D  = \frac{\sum_{ij} A_{ij}}{mn},
\end{align*}
and adjust $\hat{\mu}$ accordingly. 

\subsection{E step}\label{sec:E}
The E step concerns the computation of $J(q_1,q_2,\Phi)$, or equivalently, $\mathbb{P}_{q_1}(z_i=k)$ and $\mathbb{P}_{q_2}(w_j=l)$. At first glance, the number of terms in  both quantities grow exponentially. The next proposition shows that both $q_1$ and $q_2$ can be factorized if the other one is fixed.
\begin{proposition}\label{thm:E}
	Define
	\begin{align*}
	g_1(z_i) = &  -\sum_{j=1}^n   \theta_i \lambda_j \left ( \sum_{l=1}^L \mathbb{P}_{q_2}(w_j=l) \mu_{z_i l} \right )+ \sum_{j=1}^n  A_{ij} \left ( \sum_{l=1}^L \mathbb{P}_{q_2}(w_j=l) \log  \mu_{z_i l}  \right )+ \log \pi_{z_i}, \\
& \quad \quad \quad \quad \quad \quad \quad \quad \quad \quad \quad \quad \quad \quad \quad \quad \quad \quad \quad \quad  \quad \quad \quad \quad \quad \quad \quad  \quad \quad i=1,...,m, \\
	g_2(w_j) = &  -\sum_{i=1}^m   \theta_i \lambda_j \left ( \sum_{k=1}^K \mathbb{P}_{q_1}(z_i=k) \mu_{k w_j} \right )+ \sum_{i=1}^m  A_{ij} \left ( \sum_{k=1}^K \mathbb{P}_{q_1}(z_i=k) \log  \mu_{k w_j}  \right )+ \log \rho_{w_j}, \\
&  \quad \quad \quad \quad \quad \quad \quad \quad \quad \quad \quad \quad \quad \quad \quad \quad \quad \quad \quad \quad  \quad \quad \quad \quad \quad \quad \quad  \quad \quad j=1,...,n.
	\end{align*}
	Given $\Phi$ and $q_2$,
	\begin{align}
	\argmax_{q_1} J(q_1,q_2,\Phi)=  \prod_{i=1}^m \frac{ e^{g_1(z_i)} }{ \sum_{k=1}^K  e^{g_1(k)} }. \label{factor1}
	\end{align}
	Given $\Phi$ and $q_1$,
	\begin{align}
	\argmax_{q_2} J(q_1,q_2,\Phi) =  \prod_{j=1}^n \frac{ e^{g_2(w_j)} }{ \sum_{l=1}^L  e^{g_2(l)} }. \label{factor2}
	\end{align}
\end{proposition}
The factored form in \eqref{factor1} and \eqref{factor2} is the key reason that the variational EM is computationally feasible. It is worth emphasizing that unlike in the variational EM algorithm for symmetric networks \citep{daudin2008mixture,bickel2013asymptotic}, the factored form of $q_1$ and $q_2$ in the scenario of bipartite networks is not an assumption but a conclusion according to Proposition \ref{thm:E}. 
The factored form for the classical LBM was proved by \cite{govaert2008block}  and rediscovered by \cite{wang2021efficient}.

\subsection{Initial values}\label{sec:init}
As an iterative algorithm, the variational EM needs initial values to proceed. Our algorithm is designed to start from the M step. Therefore, we need to specify the initial values for $q_1$ and $q_2$. We use  the widely-adopted spectral clustering method \citep{ng2002spectral} on rows and columns respectively. Specifically, we carry out spectral clustering on $AA^T$ and denote the estimated row cluster labels by $\hat{{z}}^{\textnormal{init}}$. Let $q_1({z}) \propto \prod_{i=1}^m 1(z_i=\hat{z}^{\textnormal{init}}_i)$. Similarly, carry out spectral clustering on $A^TA$ and denote the estimated column labels by $\hat{{w}}^{\textnormal{init}}$. Let $q_2({w}) \propto \prod_{j=1}^n 1(w_j=\hat{w}^{\textnormal{init}}_j)$. Such  initial values $\hat{{z}}^{\textnormal{init}}$ and $\hat{{w}}^{\textnormal{init}}$ have been used in the literature \citep{Ji2020,wang2021fast}. We faithfully implement the spectral clustering algorithm described in \cite{ng2002spectral}. In particular, we normalize the embedded points (setting the norm equal to 1) before applying the $k$-means algorithm when conducting spectral clustering, as suggested in \cite{ng2002spectral} (Step 4 of the algorithm). This normalization typically aids in correcting for degree variation, in addition to the usage of the graph Laplacian. We summarize the variational EM in Algorithm \ref{alg:EM}.

\section{Asymptotic properties}\label{sec:theory}
\citet{brault2020consistency} established the consistency of the estimators for the parameters  in the classical LBM. They initially demonstrated the consistency of the maximum likelihood estimator (MLE) when cluster labels are observed. Furthermore, they showed that both the marginal likelihood and the variational approximation  are asymptotically equivalent to the complete data likelihood.  Consequently, this implies the consistency of the MLE and variational estimator of the parameters $\pi, \rho$, and $\mu$.

We adopt a different approach to studying the DC-LBM. We focus on the consistency of clustering, that is, the consistency of $q_1$ and $q_2$. 
 We prove the consistency and the convergence rate by showing that \eqref{main_obj} converges uniformly to its population version and the population has a well-separated maximizer. The proof can handle degree parameters and necessitates weaker conditions. In particular, we allow the expected graph density goes to zero as long as both the average expected row and column degrees go to infinity, which is a typical condition for label consistency for symmetric networks \citep{Bickel&Chen2009,zhao2012consistency}.

\newpage

\begin{algorithm}[ht!]
	\KwIn{$A$,  $K$, $L$;}
	Set $\hat{{z}}^{\textnormal{init}}$ (resp. $\hat{{w}}^{\textnormal{init}}$) be the outcome of spectral clustering on $AA^T$ (resp.  $A^TA$); \\
	$q_1({z}) \propto \prod_{i=1}^m 1(z_i=\hat{z}^{\textnormal{init}}_i)$;
	$q_2({w}) \propto \prod_{j=1}^n 1(w_j=\hat{w}^{\textnormal{init}}_j)$; \\
	$D=\sum_{ij} A_{ij}/(mn)$; \\	$\hat{\theta}_i=\sum_{j} A_{ij} /( n\sqrt{D} ), i=1,...,m$; $\hat{\lambda}_j=\sum_{i} A_{ij} /(m\sqrt{D}), j=1,...,n$;\\
	
	\Repeat{convergence}{
		\textbf{M step:}
		\begin{align*}
		&	 \hat{\mu}_{kl}= \frac{\sum_{i=1}^m \sum_{j=1}^n  A_{ij}  \mathbb{P}_{q_1}(z_i=k) \mathbb{P}_{q_2}(w_j=l)}{\sum_{i=1}^m \sum_{j=1}^n \hat{\theta}_i \hat{\lambda}_j \mathbb{P}_{q_1}(z_i=k) \mathbb{P}_{q_2}(w_j=l) }, \,\, k=1,...,K, l=1,...,L; \\
		&    	\hat{\pi}_k =  \frac{1}{m} \sum_{i=1}^m  \mathbb{P}_{q_1}(z_i=k), \,\, k=1,...,K; \quad \hat{\rho}_l =   \frac{1}{n} \sum_{j=1}^n  \mathbb{P}_{q_2}(w_j=l), \,\, l=1,...,L;
		\end{align*}
		\textbf{E step:}
		\begin{align*}
		& g_1(z_i) =   -\sum_{j=1}^n   \hat{\theta}_i \hat{\lambda}_j \left ( \sum_{l=1}^L \mathbb{P}_{q_2}(w_j=l) \hat{\mu}_{z_i l} \right )+ \sum_{j=1}^n  A_{ij} \left ( \sum_{l=1}^L \mathbb{P}_{q_2}(w_j=l) \log  \hat{\mu}_{z_i l}  \right ) \\
		& \quad  \quad \quad \quad + \log \hat{\pi}_{z_i}, i=1,...,m; \\
		& q_1({z}) =  \prod_{i=1}^m \frac{ e^{g_1(z_i)} }{ \sum_{k=1}^K  e^{g_1(k)} }; \\
		& g_2(w_j) =   -\sum_{i=1}^m   \hat{\theta}_i \hat{\lambda}_j \left ( \sum_{k=1}^K \mathbb{P}_{q_1}(z_i=k) \hat{\mu}_{k w_j} \right )+ \sum_{i=1}^m  A_{ij} \left ( \sum_{k=1}^K \mathbb{P}_{q_1}(z_i=k) \log  \hat{\mu}_{k w_j}  \right ) \\
		& \quad \quad \quad \quad + \log \hat{\rho}_{w_j}, j=1,...,n;\\
		&  q_2({w}) = \prod_{j=1}^n \frac{ e^{g_2(w_j)} }{ \sum_{l=1}^L  e^{g_2(l)} };
		\end{align*}
	}
	\KwOut{$q_1$, $q_2$.}
	\caption{Variational EM algorithm for the DC-LBM} \label{alg:EM}
\end{algorithm}

\newpage 

Let ${\pi}^*,{\rho}^*$ be the true prior probabilities and  ${z}^*,{w}^*$ be the true row and column labels. According to the model assumption, the conditional expectation of $A_{ij}$ has the form
\begin{align}
E[A_{ij}| z_i^*,w_j^*]=\theta_i\lambda_j \mu_{z^*_iw^*_j}. \label{uncanonical}
\end{align}
The parameters ${\theta}, {\lambda}$ and $\mu$ are clearly not identifiable. We therefore introduce the following canonical form for ${\theta}, {\lambda}$ and $\mu$:
\begin{align}
\theta^*_i & =  \frac{\frac{1}{n}\sum_{j=1}^n E[A_{ij}| z_i^*,w_j^*]}{(\frac{1}{mn} \sum_{i=1}^m\sum_{j=1}^n E[A_{ij}|z_i^*,w_j^*] )^{1/2}}, \,\, i=1,...,m, \nonumber \\
\lambda^*_j & =  \frac{\frac{1}{m}\sum_{i=1}^m E[A_{ij}| z_i^*,w_j^*]}{(\frac{1}{mn} \sum_{i=1}^m\sum_{j=1}^n E[A_{ij}|z_i^*,w_j^*] )^{1/2}}, \,\, j=1,...,n,  \nonumber \\
\mu^*_{kl} & =  \frac{E[A_{ij}| z_i^*,w_j^*]}{\theta^*_i  \lambda^*_j}, \,\, \textnormal{for some $i,j$ with $z_i^*=k,w_j^*=l$}, \,\, k=1,...,K,l=1,...,L. \label{cano_mu}
\end{align}
We need to show that $\mu^*_{kl}$ are well defined; in other word, its value depends on row and column labels but not the row and column indices, which is given in the following proposition.
\begin{proposition}\label{thm:well_defined}
	$\mu^*_{kl}$ defined in \eqref{cano_mu} depends on $i$ and $j$ only through their cluster labels $z_i^*=k$ and $w_j^*=l$.
\end{proposition}
It is worth mentioning that, among the many possibilities for choosing canonical parameters, our selection is particularly convenient for the following reasons: The mean parameters $\mu^*_{kl}\asymp 1$ under this definition. Thus, the density of the network is fully characterized by ${\theta_i^*}$ and ${\lambda_j^*}$ (Proposition 4).  In the meantime, $\theta_i^*$ and $\lambda_j^*$ can be  directly and consistently estimated  by (functions of) row and column degrees, without involving any unknown scale factor. Those properties facilitate the theoretical analysis (Theorem 6 and Theorem 9).


We make the following assumptions on the parameters throughout the theoretical analysis:
\begin{itemize}
	\item[$H_1:$] $\pi_{\textnormal{min}}\leq \pi^*_k \leq \pi_{\textnormal{max}}\,\,(k=1,...,K)$  and $\rho_{\textnormal{min}}\leq \rho^*_l \leq \rho_{\textnormal{max}}\,\,(l=1,...,L)$, where $\pi_{\textnormal{min}}$, $\pi_{\textnormal{max}}$, $\rho_{\textnormal{min}}$ and $\rho_{\textnormal{max}}$ are positive constants. 
 Furthermore, $\tilde{\pi}_{\textnormal{min}}\leq (1/m) \sum_i 1(z^*_i=k) \leq \tilde{\pi}_{\textnormal{max}}\,\,(k=1,...,K)$  and $\tilde{\rho}_{\textnormal{min}}\leq (1/n)\sum_j 1(w^*_j=l)  \leq \tilde{\rho}_{\textnormal{max}}\,\,(l=1,...,L)$, where $\tilde{\pi}_{\textnormal{min}}$, $\tilde{\pi}_{\textnormal{max}}$, $\tilde{\rho}_{\textnormal{min}}$ and $\tilde{\rho}_{\textnormal{max}}$ are positive constants.
	
	\item[$H_2:$]  $E[A_{ij}| z_i^*,w_j^*]=r_{mn} E_{ij}\,\,(i=1,...,m,\,j=1,...,n)$ where $(mn r_{mn} )/(m+n  ) \rightarrow \infty$ as $m,n\rightarrow \infty$, and $ 0< E_{\textnormal{min}} \leq E_{ij} \leq E_{\textnormal{max}} < \infty$, where $E_{\textnormal{min}}$ and $E_{\textnormal{max}}$ are positive constants.
	\item[$H_3:$] Each row and each column of $\mu^*$ is unique. That is, there do not exist two rows $k$ and $k'$ such that $\mu^*_{kl}=\mu^*_{k'l}$ for all $l$, and there do not exist two columns $l$ and $l'$ such that $\mu^*_{kl}=\mu^*_{kl'}$ for all $k$.
\end{itemize}

All assumptions above are standard.
Assumption $H_1$  ensures that no cluster size is too small. The second part of $H_1$ in fact automatically holds with high probability given the first part, which can be proved by applying Hoeffding's inequality (see Proposition 4.2 in \cite{brault2020consistency} for example). Here we directly assume the condition for simplicity. Assumption $H_2$ is an analogue of a typical assumption on graph density in many works from the community detection literature \citep{Bickel&Chen2009,zhao2012consistency}. Specifically, the factor $r_{mn}$ measures the rate of the expected graph density decay. Because $(mn r_{mn})/(m+n) \geq (mn r_{mn})/(2\max(m,n))$, $(mn r_{mn})/(m+n) \rightarrow \infty$ if and only if $nr_{mn}\rightarrow \infty$ and $m r_{mn}\rightarrow \infty$. That is, the average expected row and column degrees go to infinity.
Assumption $H_2$ has the following implications on the canonical parameters.
\begin{proposition}
	Under Assumption $H_2$, for $i=1,...,m$, $\theta_{\textnormal{min}} \sqrt{r_{mn}} \leq \theta_i^* \leq \theta_{\textnormal{max}}\sqrt{r_{mn}}$, where $\theta_{\textnormal{min}},\theta_{\textnormal{max}}$ are positive constants. Similarly, $\lambda_{\textnormal{min}} \sqrt{r_{mn}} \leq \lambda^*_j \leq \lambda_{\textnormal{max}}\sqrt{r_{mn}} $ for $j=1,...,n$, where $\lambda_{\textnormal{min}},\lambda_{\textnormal{max}}$ are positive constants.
	Finally, $\mu_{\textnormal{min}} \leq  \mu^*_{kl} \leq \mu_{\textnormal{max}}$, where $\mu_{\textnormal{min}}$ and $\mu_{\textnormal{max}}$ are positive constants.
\end{proposition}
The proof is straightforward and hence is omitted.

Assumption $H_3$ has the same form of $H_3$ in \cite{brault2020consistency}, which ensures $\mu^*$ is identifiable up to a
permutation of the row  and column labels. In the DC-LBM, this assumption is satisfied if no two rows (columns) of any  $\mu$ that gives \eqref{uncanonical} are proportional to each other. The next proposition elaborates on the details.
\begin{proposition}\label{thm:iden_cond}
	$H_3$ holds if and only if for any parametrization $({\theta},{\lambda},\mu)$ satisfying \eqref{uncanonical}, there do not exist two rows $k$ and $k'$ such that
	$\mu_{kl}=\mu_{k'l}a_{kk'}$ for all $l$ and there do not exist two columns $l$ and $l'$ such that $\mu_{kl}=\mu_{kl'}b_{ll'}$ for all $k$.
\end{proposition}

Our goal is to study the property of the variational approximation $J(q_1,q_2,\Phi)$. Proposition \ref{thm:E} shows that  $q_1$ (resp. $q_2$) can be factorized if $q_2$ (resp. $q_1$) is given. We therefore study $J(q_1,q_2,\Phi)$ under the constraint that $q_1$ and $q_2$ are product measures, which can therefore be represented as matrices.
Let $q^{{z}}=[q^{{z}}_{ik}]_{m \times K}$ and $q^{{w}}=[q^{{w}}_{jl}]_{n \times L}$ be $m \times K$ and $n \times L$ matrices with $q^{{z}}_{ik}=\mathbb{P}_{q_1}(z_i=k)$ and $q^{{w}}_{jl}=\mathbb{P}_{q_2}(w_j=l)$.  We rewrite $J(q_1,q_2,\Phi)$ in terms of $q^{{z}},q^{{w}}$:
\begin{align*}
J(q^{{z}},q^{{w}},\Phi)   = \,\, &   -\sum_{i=1}^m \sum_{j=1}^n  \theta_i \lambda_j  \left (\sum_{k=1}^K \sum_{l=1}^L q^{{z}}_{ik} q^{{w}}_{jl} \mu_{k l} \right )  + \sum_{i=1}^m \sum_{j=1}^n   A_{ij} \left (\sum_{k=1}^K \sum_{l=1}^L q^{{z}}_{ik} q^{{w}}_{jl} \log  \mu_{kl}  \right )  \\
& + \sum_{i=1}^m \sum_{j=1}^n  A_{ij} (\log  \theta_i+\log \lambda_j)   +  \sum_{i=1}^m  \left ( \sum_{k=1}^K  q^{{z}}_{ik} \log \pi_k \right ) + \sum_{j=1}^n \left ( \sum_{l=1}^L  q^{{w}}_{jl}  \log \rho_l \right ) \\
&  - \sum_{i=1}^m  \sum_{k=1}^K q^{{z}}_{ik} \log q^{{z}}_{ik} -\sum_{j=1}^n \sum_{l=1}^L q^{{w}}_{jl} \log q^{{w}}_{jl}.
\end{align*}
Furthermore,  we replace $\theta_i$ and $\lambda_j$ in $J(q^{{z}},q^{{w}},\Phi)$  by the estimators derived in Proposition \ref{thm:stationary} since they remain unchanged throughout the algorithm:
\begin{align}
& \hat{\theta}_i  = \frac{d_i^r}{\sqrt{D}}, \,\, i=1,...,m, \,\, \hat{\lambda}_j  = \frac{d_j^c}{\sqrt{D}}, \,\, j=1,...,n, \,\, \textnormal{where }  D  = \frac{\sum_{ij} A_{ij}}{mn}. \label{degree}
\end{align}
We exclude $\theta_i$ and $\lambda_j$  from $\Phi$ thereafter, and omit the term $\sum_{i=1}^m \sum_{j=1}^n  A_{ij} (\log  \theta_i+\log \lambda_j)$ in $J$ because it does not affect the estimation of $q^{{z}}$ and $q^{{w}}$ when $\hat{{\theta}}$ and $\hat{{\lambda}}$ are fixed.
Denote the new criterion function by
\begin{align}
& \hat{J}(q^{{z}},q^{{w}},\Phi)  =    -\sum_{i=1}^m \sum_{j=1}^n  \hat{\theta}_i \hat{\lambda}_j  \left (\sum_{k=1}^K \sum_{l=1}^L q^{{z}}_{ik} q^{{w}}_{jl} \mu_{k l} \right )  + \sum_{i=1}^m \sum_{j=1}^n   A_{ij} \left (\sum_{k=1}^K \sum_{l=1}^L q^{{z}}_{ik} q^{{w}}_{jl} \log  \mu_{kl}  \right ) \nonumber  \\
&\quad  +  \sum_{i=1}^m  \left ( \sum_{k=1}^K  q^{{z}}_{ik} \log \pi_k \right )  + \sum_{j=1}^n \left ( \sum_{l=1}^L  q^{{w}}_{jl}  \log \rho_l \right )  - \sum_{i=1}^m  \sum_{k=1}^K q^{{z}}_{ik} \log q^{{z}}_{ik} -\sum_{j=1}^n \sum_{l=1}^L q^{{w}}_{jl} \log q^{{w}}_{jl}. \label{criterion}
\end{align}
The next theorem shows $\hat{J}(q^{{z}},q^{{w}},\Phi)$ uniformly converges to its ``population version'' omitting lower order terms
\begin{align*}
\bar{J}(q^{{z}},q^{{w}},\mu)  = \,\, &   -\sum_{i=1}^m \sum_{j=1}^n  \theta^*_i \lambda^*_j  \left (\sum_{k=1}^K \sum_{l=1}^L q^{{z}}_{ik} q^{{w}}_{jl} \mu_{k l} \right ) \\
& + \sum_{i=1}^m \sum_{j=1}^n    E [A_{ij} |{z}^*,{w}^*] \left (\sum_{k=1}^K \sum_{l=1}^L q^{{z}}_{ik} q^{{w}}_{jl} \log  \mu_{kl}  \right ).
\end{align*}
\begin{theorem}\label{thm:uniform}
	If $(mn r_{mn} \epsilon^2)/(m+n) \rightarrow \infty$  as $m,n\rightarrow \infty$, for $\epsilon$ being a positive constant or $o(1)$, then, under $H_1$ and $H_2$, we have
	\begin{align*}
	\mathbb{P} \left ( \left .  \max_{q^{{z}}\in \mathcal{C}_{{z}},q^{{w}}\in \mathcal{C}_{{w}}, {\pi}\in \mathcal{C}_{{\pi}}, {\rho}\in \mathcal{C}_{{\rho}},\mu \in \mathcal{C}_{\mu}  } \left |\hat{J}(q^{{z}} ,q^{{w}},\Phi)-\bar{J}(q^{{z}},q^{{w}},\mu)\right | \geq mn r_{mn} \epsilon
	\right | {z}^*,{w}^* \right )\rightarrow  0,
	\end{align*}
	where $\mathcal{C}_{{z}}$, $\mathcal{C}_{{w}}$, $\mathcal{C}_{{\pi}}$, $\mathcal{C}_{{\rho}}$, and  $\mathcal{C}_{\mu}$ are the (compact) domains for the corresponding parameters. Specifically, $\mathcal{C}_{{z}}=\{ q^{{z}}:  q^{{z}} \in \mathbb{R}^{m \times K},\, q^{{z}}_{ik}\in [0,1],\, \sum_{k=1}^K q^{{z}}_{ik}=1\}$, $\mathcal{C}_{{w}}=\{ q^{{w}}:  q^{{w}} \in \mathbb{R}^{n \times L},\, q^{{w}}_{jl}\in [0,1],\, \sum_{l=1}^L q^{{w}}_{jl}=1\}$, $\mathcal{C}_{{\pi}}= \{ {\pi}: {\pi} \in \mathbb{R}^{K},\, \pi_k \in [\pi_{\textnormal{min}},\pi_{\textnormal{max}}],\, \sum_{k=1}^K\pi_k=1   \}$, $\mathcal{C}_{{\rho}}= \{ {\rho}: {\rho} \in \mathbb{R}^{L},\, \rho_l \in [\rho_{\textnormal{min}},\,\rho_{\textnormal{max}}],\, \sum_{l=1}^L\rho_l=1   \}$, and $\mathcal{C}_{\mu}=\{ \mu:\mu\in \mathbb{R}^{K \times L},\, \mu_{\textnormal{min}} \leq  \mu_{kl} \leq \mu_{\textnormal{max}} \}$.
\end{theorem}

Next we state a result that the true labels (up to a permutation) are a well-separated maximizer of $\bar{J}(q^{{z}},q^{{w}},\mu)$. We give more definitions before proceeding.

%
\begin{definition}[Soft confusion matrix]
	For any row label assignment matrices $q^{{z}}$ and $\tilde{q}^{{z}}$, let
	\begin{align*}
	\mathbb{R}_{ kk'}(q^{{z}},\tilde{q}^{{z}}) & =\frac{1}{m} \sum_{i=1}^m q^{{z}}_{ik} \tilde{q}^{{z}}_{ik'}.
	\end{align*}
	In particular, the confusion matrix for the true row label ${z}^*$ and $q^{{z}}$ is
	\begin{align*}
	\mathbb{R}_{ kk'}({1}^{{z}^*},q^{{z}}) & =\frac{1}{m} \sum_{i=1}^m  {1}^{{z}^*}_{ik} q^{{z}}_{ik'},
	\end{align*}
	where ${1}^{{z}^*}_{ik}=1(z^*_i=k)$.
	
	Similarly, for any column label assignments, let
	\begin{align*}
	\mathbb{R}_{ ll'}(q^{{w}},\tilde{q}^{{w}})  =\frac{1}{n} \sum_{j=1}^n q^{{w}}_{jl} \tilde{q}^{{w}}_{jl'}, \,\, \mathbb{R}_{ ll'}({1}^{{w}^*},q^{{w}})  =\frac{1}{n} \sum_{j=1}^n {1}^{{w}^*}_{jl} q^{{w}}_{jl'},
	\end{align*}
	where ${1}^{{w}^*}_{jl}=1(w^*_j=l)$.
\end{definition}
The soft confusion matrices defined above generalize the confusion matrix for comparing two hard label assignments to the case of comparing two probability matrices. Let $S_K$ ($S_L$) be the set of permutations on $\{1,...,K\}$ ($\{1,...,L\}$). Taking permutations into account, the misclassification rates for row clusters and column clusters are defined as
\begin{align*}
M_\textnormal{row}(q^{{z}}) & = \min_{s \in S_K} \left ( 1-   \sum_{k'=1}^K  \mathbb{R}_{ s(k'),k'}({1}^{{z}^*},q^{{z}}) \right),  \\
M_\textnormal{col}(q^{{w}}) & = \min_{t \in S_L} \left ( 1-   \sum_{l'=1}^L  \mathbb{R}_{ t(l'),l'}({1}^{{w}^*},q^{{w}}) \right).
\end{align*}

The following theorem shows that $\bar{J}({1}^{{z}^*},{1}^{{w}^*},\mu^*)-\bar{J}(q^{{z}},q^{{w}},\mu)$ is bounded below by $M_\textnormal{row}(q^{{z}})$ and $M_\textnormal{col}(q^{{w}})$.
\begin{theorem}\label{thm:separability}
	For all $q^{{z}}\in \mathcal{C}_{{z}}$, $q^{{w}}\in \mathcal{C}_{{w}}$, and $\mu \in \mathcal{C}_\mu$, under $H_1$ and $H_3$, we have
	\begin{align*}
	\bar{J}({1}^{{z}^*},{1}^{{w}^*},\mu^*)-\bar{J}(q^{{z}},q^{{w}},\mu) & \geq C_1 mn r_{mn}  M_\textnormal{row}(q^{{z}}), \\
    \bar{J}({1}^{{z}^*},{1}^{{w}^*},\mu^*)-\bar{J}(q^{{z}},q^{{w}},\mu) & \geq C_2 mn r_{mn}  M_\textnormal{col}(q^{{w}}).
	\end{align*}
\end{theorem}

Finally, we give the rate of convergence of the misclassification rate, which implies label consistency.
Let $\hat{\Phi}=(\hat{\mu},\hat{{\pi}},\hat{{\rho}})$ be a maximizer of $J(q^{{z}},q^{{w}},\Phi)$,
that is,
\begin{align*}
(\hat{q}^{{z}},\hat{q}^{{w}},\hat{\Phi}) = \argmax_{q^{{z}}\in \mathcal{C}_{{z}},q^{{w}}\in \mathcal{C}_{{w}}, \mu \in \mathcal{C}_{\mu}, {\pi}\in \mathcal{C}_{{\pi}}, {\rho}\in \mathcal{C}_{{\rho}}  } \hat{J}(q^{{z}} ,q^{{w}},\Phi).
\end{align*}

\begin{theorem} \label{thm:consistency}
	Assume $H_1,H_2$ and $H_3$. If $(mn r_{mn} )/(m+n  ) \rightarrow \infty$ as $m,n \rightarrow \infty$, then for all positive constant $\delta$, we have
\begin{align*}
M_\textnormal{row}(\hat{q}^{{z}})=o_p \left ( \left ( \frac{mnr_{mn}}{m+n}\right)^{-1/2+\delta} \right), \,\, M_\textnormal{col}(\hat{q}^{{w}})=o_p \left ( \left ( \frac{mnr_{mn}}{m+n}\right)^{-1/2+\delta} \right).
\end{align*}
\end{theorem}

Finally, we provide the rate of convergence of misclassification for networks with binary edges, parallel to the result in Theorem \ref{thm:consistency}. Note that the closed-form solutions $\{\hat{\theta}_i\}$ and $\{\hat{\lambda}_i\}$, and thus the algorithm, rely on the form of the Poisson distribution. However, if we apply the same estimating procedure to a network with edges following Bernoulli distributions, we can establish the same convergence rate as in Theorem \ref{thm:consistency}. Specifically, we assume that given labels, $A_{ij}$'s independently follow Bernoulli distributions with $E[A_{ij}| z_i^*,w_j^*]=\theta_i\lambda_j \mu_{z^*_iw^*_j}$. But we analyze the estimator from the Poisson model---that is, $\hat{\theta}_i$ and $\hat{\lambda}_j$ are computed by \eqref{degree} and $\hat{\Phi}=(\hat{\mu},\hat{{\pi}},\hat{{\rho}})$ is still a maximizer of $J(q^{{z}},q^{{w}},\Phi)$, defined by \eqref{criterion}. We obtain the following result.
\begin{theorem} \label{thm:consistency_ber}
	Let $A_{ij}$'s  independently follow Bernoulli distributions with $E[A_{ij}| z_i^*,w_j^*]=\theta_i\lambda_j \mu_{z^*_iw^*_j}$, given $z^*$ and $w^*$. Assume the canonical parameters satisfy $H_1,H_2$ and $H_3$. If $(mn r_{mn} )/(m+n  ) \rightarrow \infty$ as $m,n \rightarrow \infty$, then for all positive constant $\delta$, we have
\begin{align*}
M_\textnormal{row}(\hat{q}^{{z}})=o_p \left ( \left ( \frac{mnr_{mn}}{m+n}\right)^{-1/2+\delta} \right), \,\, M_\textnormal{col}(\hat{q}^{{w}})=o_p \left ( \left ( \frac{mnr_{mn}}{m+n}\right)^{-1/2+\delta} \right).
\end{align*}
\end{theorem}
The proof is similar to that of Theorem \ref{thm:consistency}. Note that all results, except for Theorem \ref{thm:uniform}, only concern the mean parameters and therefore hold true under the Bernoulli model. The proof of Theorem \ref{thm:uniform} relies on a concentration inequality of Poisson variables \citep{Canonne_Poisson}, which can be replaced by the Bernstein inequality of Bernoulli variables. We refer the readers to the appendix for details. 

\section{Simulation studies}\label{sec:simu}
In this section, we compare the proposed variational EM algorithm for the DC-LBM to two other methods, the  profile likelihood based biclustering
method  \citep{flynn2020profile} and spectral clustering \citep{ng2002spectral}. The profile likelihood based biclustering extends the classical SBM to  biclustering. The method assumes that $\{ A_{ij} \}$ are sampled from distributions in an exponential family, which gives a flexible choice, such as Bernoulli, Poisson and Gaussian. However, the likelihood does not incorporate degree parameters. The model is identical to the classical LBM when assuming the Bernoulli distribution on $\{ A_{ij} \}$. The profile likelihood based biclustering method treats cluster labels to be unknown fixed parameters and a local search technique based on the Kernighan-Lin heuristic \citep{kernighan1970efficient} was applied to search the optimal row and column partitions. To reduce the possibility of the  algorithm finding a local optimum, we use 30 random initial partitions in all simulation settings.
We apply spectral clustering to biclustering in the same manner as described in Section \ref{sec:init}---that is, carry out spectral clustering on $AA^T$ to find the row labels and on $A^TA$ to find the column labels.

We first evaluate the performance of the algorithms under the correctly-specified model for our proposed method---that is, the DC-LBM with the Poisson distribution. We simulate networks with the number of rows $m=800$, the number of columns $n=1000$, the number of row clusters $K=3$, and the number of column clusters $L=4$. Row cluster labels $\{z_i \}$ are independently generated from $\textnormal{Multi}(1,{\pi}=(1/3,1/3,1/3)^{T})$, and similarly, column cluster labels $\{w_j\}$ are independently generated from $\textnormal{Multi}(1,{\rho}=(1/4,1/4,1/4,1/4)^{T})$. Degree parameters $\{\theta_i\}$ and $\{\lambda_j \}$ are independently generated from $\textnormal{Uniform}(0.5,1.5)$. For clarification, we will not use the prior information on $\{\theta_i\}$ and $\{\lambda_j \}$ in the algorithm. That is, we treat them as unknown fixed parameters.
Furthermore, we set
\begin{align}
\mu= r \begin{pmatrix}
0.15 & 0.05 & 0.05 & 0.06 \\
0.05 & 0.15 & 0.05 & 0.08 \\
0.05 & 0.05 & 0.15 & 0.10 \\
\end{pmatrix}, \label{mu_simulation}
\end{align}
where $r$ varies from 0.4, 0.6, 0.8 to 1, which controls the graph density.

We measure the accuracy of clustering by the adjusted Rand index  \citep{vinh2010information}, which is a  widely-used  measure  for  comparing  two  partitions.  The zero value of the index corresponds to two independent partitions, and higher  values  indicate  better  agreement. All reported adjusted Rand index values in the figures below were based on 200 replicates. The error bars represent the range of plus or minus one standard deviation. To enhance readability, we jitter the error bars along the x-axis. However, all methods correspond to the same $r$ values, namely $r=0.4, 0.6, 0.8$, and 1.

\begin{figure}[!ht]
	\twoImages{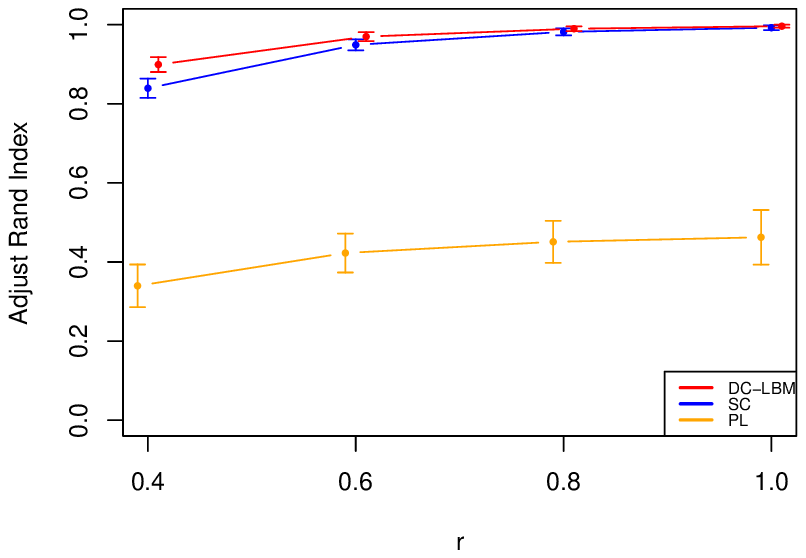}{7cm}{(a) Row communities}{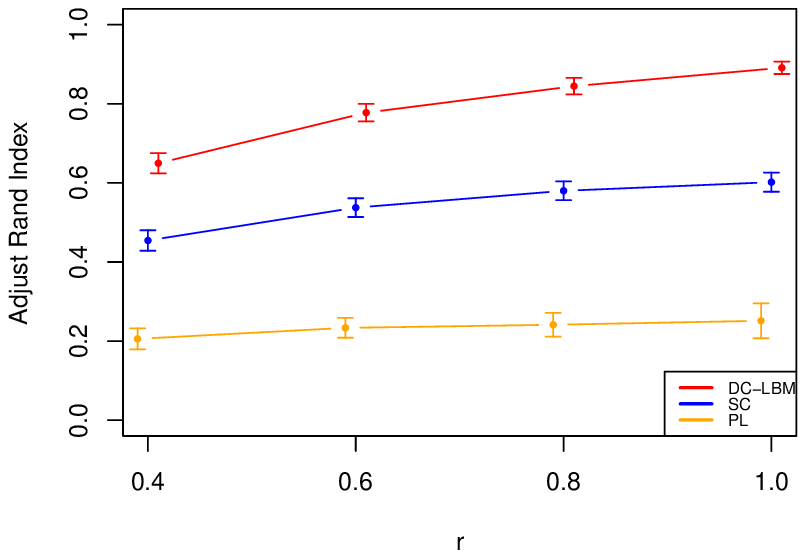}{7cm}{(b) Column communities}
	\caption{Performance of three algorithms under the DC-LBM with the Poisson distribution. $r$: the graph density factor  in \eqref{mu_simulation}. Left panel: detection of row clusters. Right panel: detection of column clusters. SC: spectral clustering. PL: profile likelihood based biclustering. }
	\label{fig:deg_poisson}
\end{figure}

Figure \ref{fig:deg_poisson} shows the performance of three algorithms for detecting row and column clusters under the DC-LBM with the Poisson distribution.  Our first observation is simply that the adjusted Rand index values for all methods improve as the graph density increases, which is in line with common sense in the community detection literature. Second, profile likelihood based biclustering gives the lowest adjusted Rand index values because the degree parameters are not considered by this method. Third, the performance of spectral clustering is between DC-LBM and profile likelihood based biclustering,  because although not being a model-based approach, spectral clustering implicitly takes degree variation into account. The normalized Laplacian matrix includes the diagonal matrix whose elements are the row sums of $AA^T$ (resp. $A^TA$). Moreover, the spectral clustering algorithm in \cite{ng2002spectral} renormalizes each row of the matrix whose columns are the $K$ (resp. $L$) top eigenvectors of  $AA^T$ (resp. $A^TA$). This step alleviates the effect of degree variation.
Lastly, the reason the improvement of the proposed method on row clustering is less substantial than the improvement on column clustering is that row clustering is easier in our setup, and therefore, has less room for improvement. It is noteworthy that all methods achieve better accuracy for row clustering than for column clustering in the simulations. The clustering problem on rows is easier according to the simulation setup for the following reasons: Firstly, the rows are grouped into a smaller number of clusters. Secondly, the estimated cluster label of row $i$ (resp. column $j$) is mainly determined by the row vector $A_{i\cdot}$ (resp. the column vector $A_{\cdot j}$), and a vector that contains more entries (in our case, the row vector) provides a clearer pattern. This is analogous to cluster analysis in a high-dimensional Euclidean space, where a large number of features aid in distinguish the clusters among the observations.

\begin{figure}[!ht]
	\twoImages{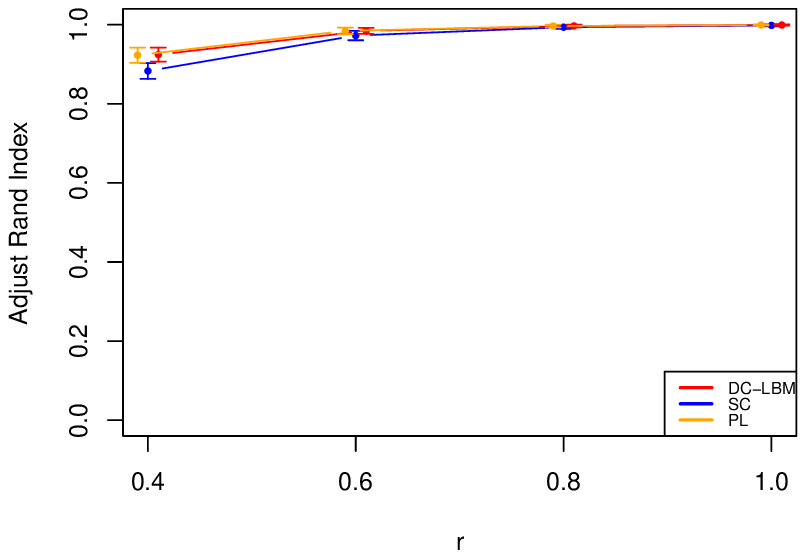}{7cm}{(a) Row communities}{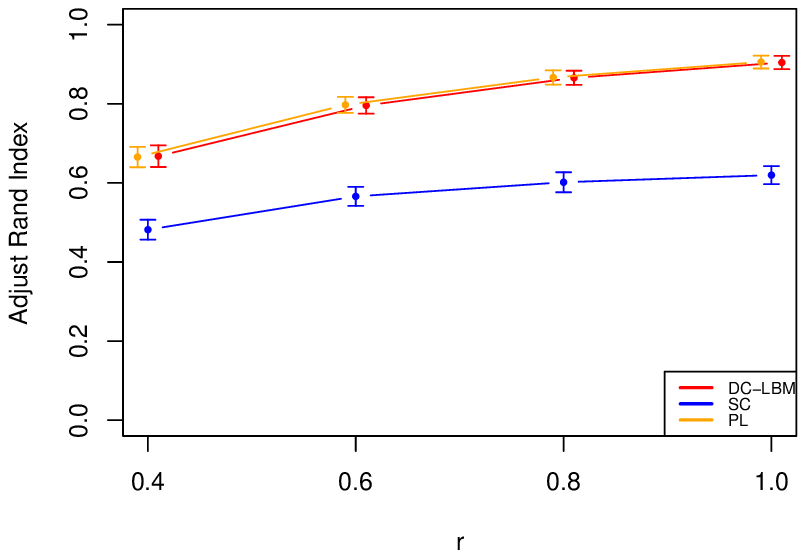}{7cm}{(b) Column communities}
	\caption{Performance of three algorithms under the classical LBM with the Poisson distribution. $r$: the graph density factor  in \eqref{mu_simulation}. The red curve and green curve are almost identical. Left panel: detection of row clusters. Right panel: detection of column clusters. SC: spectral clustering. PL: profile likelihood based biclustering. }
	\label{fig:nodeg_poisson}
\end{figure}

Our second simulation investigates how well the variational EM algorithm designed for the DC-LBM performs if the true model has equal degree parameters, that is, under the classical LBM with Poisson distribution. The model for this simulation is identical to the previous setup except that $\theta_i\equiv 1, \lambda_j\equiv 1, i=1,...,m,j=1,...,n$. From Figure \ref{fig:nodeg_poisson}, we can see that the adjusted Rand index values of the DC-LBM and of profile likelihood based biclustering are almost identical, which means  that DC-LBM loses very little efficiency when introducing the extra parameters $\{\theta_i \}$ and $\{\lambda_j \}$, and can be safely used even if the true model is the classical LBM.

\begin{figure}[!ht]
	\twoImages{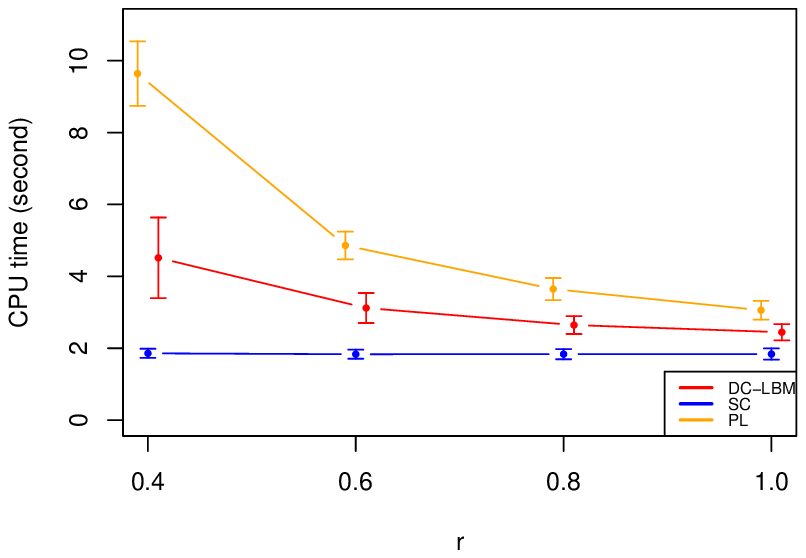}{7cm}{(a) LBM, Poisson}{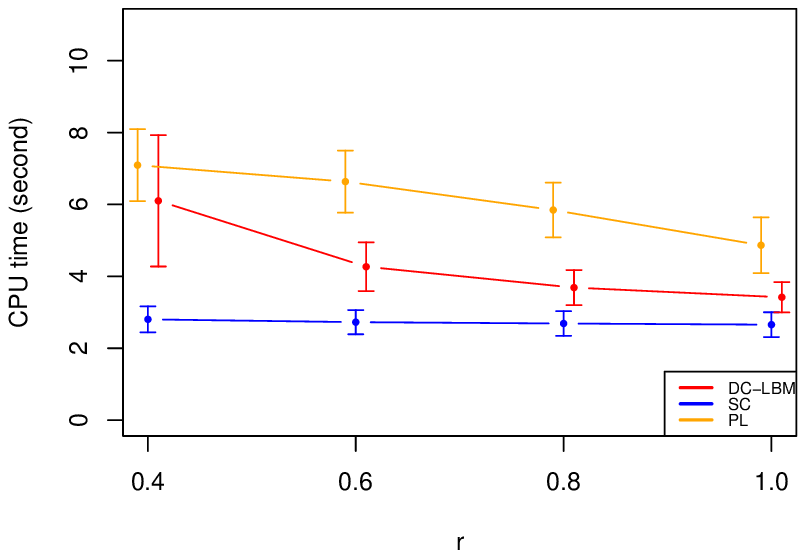}{7cm}{(b) DC-LBM, Poisson}
 \caption{CPU time for three algorithms under Poisson models. $r$: the graph density factor  in \eqref{mu_simulation}. Left panel:  the true model is the classical LBM with the Poisson distribution. Right panel: the true model is the  DC-LBM with the Poisson distribution.  } \label{fig:time_poisson}
\end{figure}

Additionally, we compare the CPU time for the three algorithms under both the classical LBM and the DC-LBM. The experiments are conducted on a high-performance computing cluster with fifth-generation Intel Core processors, and the results are reported in Figure \ref{fig:time_poisson}.  The recorded CPU time for the variational EM algorithm includes the time for spectral clustering, as it serves as the initial step in Algorithm \ref{alg:EM}. It can be seen that the variational EM algorithm costs less time than the profile likelihood based biclustering in the above simulations. Another notable pattern is that the running time for both the variational EM algorithm and the profile likelihood-based biclustering decreases as the graph density increases. This is because a network with a clearer community structure makes convergence easier for both algorithms.

\begin{figure}[!ht]
	\twoImages{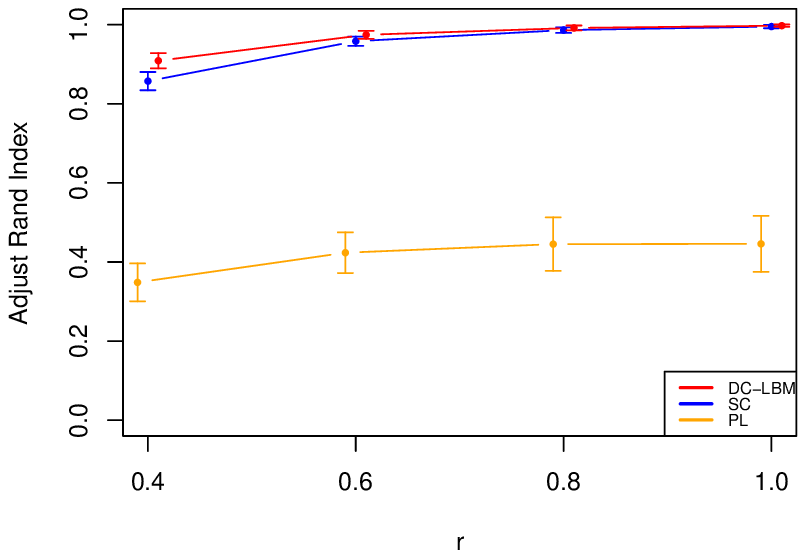}{8cm}{(a) Row communities}{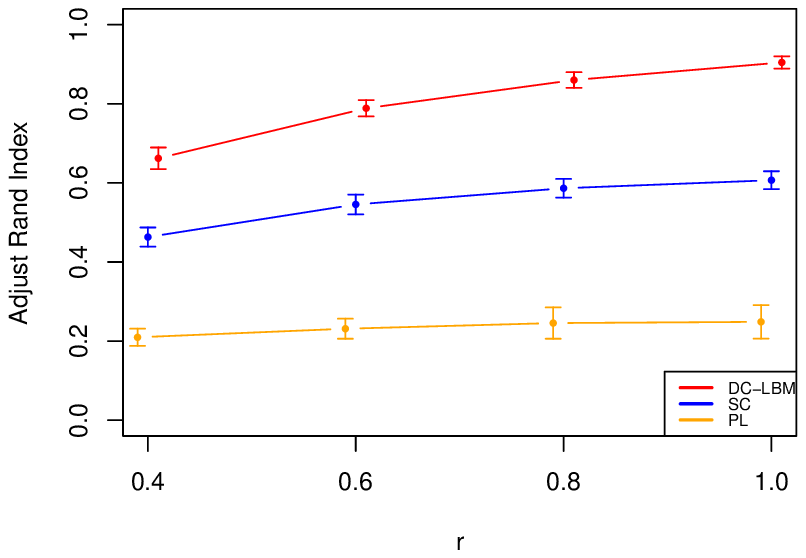}{8cm}{(b) Column communities}
	\caption{Performance of three algorithms under the DC-LBM with the Bernoulli distribution. $r$: the graph density factor  in \eqref{mu_simulation}. Left panel: detection of row clusters. Right panel: detection of column clusters. SC: spectral clustering. PL: profile likelihood based biclustering.}
	\label{fig:deg_bernoulli}
\end{figure}

\begin{figure}[!ht]
	\twoImages{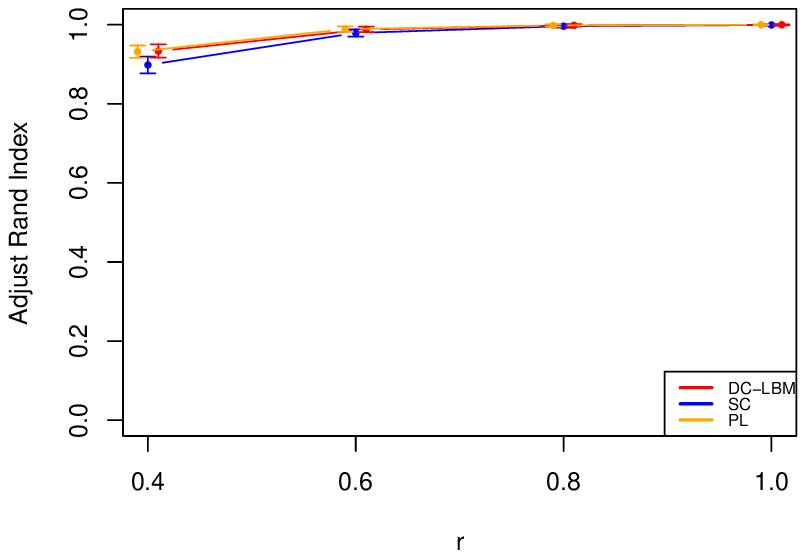}{8cm}{(a) Row communities}{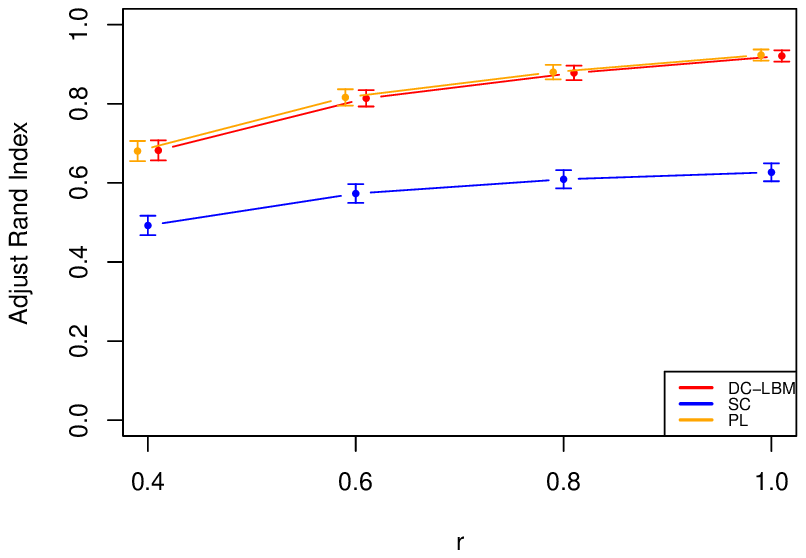}{8cm}{(b) Column communities}
	\caption{Performance of three algorithms under the classical LBM with the Bernoulli distribution. $r$: the graph density factor  in \eqref{mu_simulation}. Left panel: detection of row clusters. Right panel: detection of column clusters. SC: spectral clustering. PL: profile likelihood based biclustering.}
	\label{fig:nodeg_bernoulli}
\end{figure}

\begin{figure}[!ht]
	\twoImages{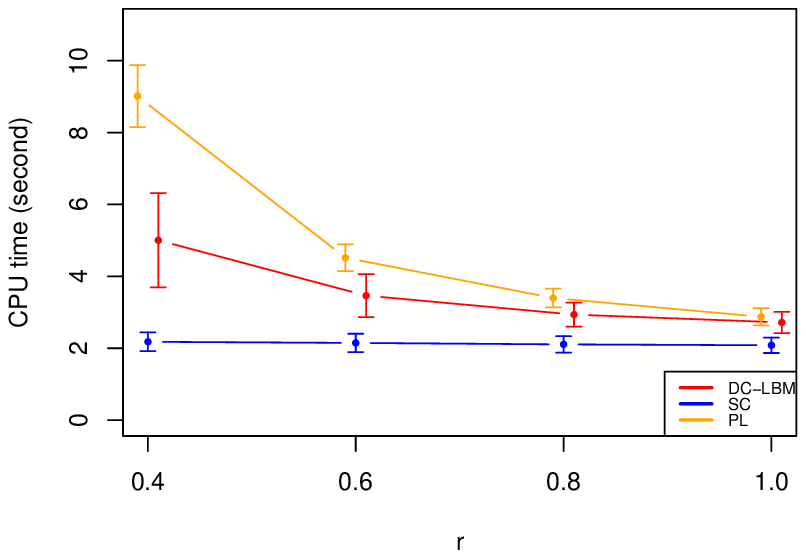}{7cm}{(a) LBM, Bernoulli}{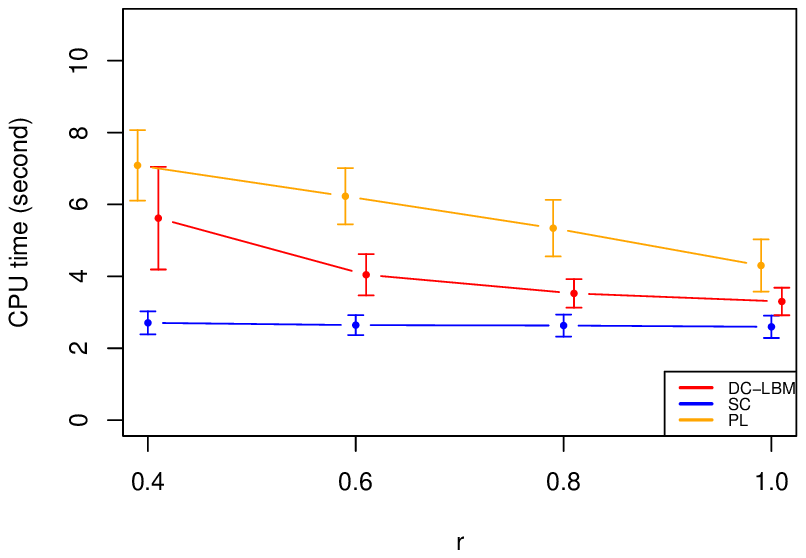}{7cm}{(b) DC-LBM, Bernoulli}
 \caption{CPU time for three algorithms under Bernoulli models. $r$: the graph density factor  in \eqref{mu_simulation}. Left panel:  the true model is the classical LBM with the Bernoulli distribution. Right panel: the true model is the  DC-LBM with the Bernoulli distribution.  } \label{fig:time_Bernoulli}
\end{figure}

Now, we evaluate the performance of the variational EM algorithm when the likelihood of $A$ is misspecified. We assume a Poisson distribution on $A_{ij}$ given the labels in our method, which is mainly due to the consideration of computation. In particular, $\{\hat{\theta}_i\}$ and $\{\hat{\lambda}_j\}$ in the M step have  a closed-form solution under the Poisson assumption, which is row and column degrees, respectively (Proposition 1). On the contrary, many real-world networks are unweighted  graphs. Therefore, we investigate the behavior of the Poisson model when $\{A_{ij}\}$ are actually Bernoulli variables. 
We carry out the two aforementioned simulations  under the Bernoulli model. That is, the parameter settings are identical to the previous two simulations except that conditional on the cluster labels, $\{A_{ij}\}$ follow $\textnormal{Ber}(\theta_i \lambda_j\mu_{z_i w_j})$. The algorithms for DC-LBM and spectral clustering are the same as before and profile likelihood based biclustering is implemented with the Bernoulli link function specified. Figure \ref{fig:deg_bernoulli} and Figure \ref{fig:nodeg_bernoulli} are identical to Figure 1 and Figure 2, which implies that the performance of the variational EM algorithm with the Poisson distribution is almost not affected by whether the true underlying distribution is Bernoulli or Poisson. Additionally, we report the CPU time for the three algorithms under both the classical LBM and the DC-LBM with the Bernoulli distribution in Figure \ref{fig:time_Bernoulli}. The pattern is again similar to that observed in the Poisson case (Figure \ref{fig:time_poisson}).

\begin{figure}[!ht]
\begin{center}
 \includegraphics[width=8cm]{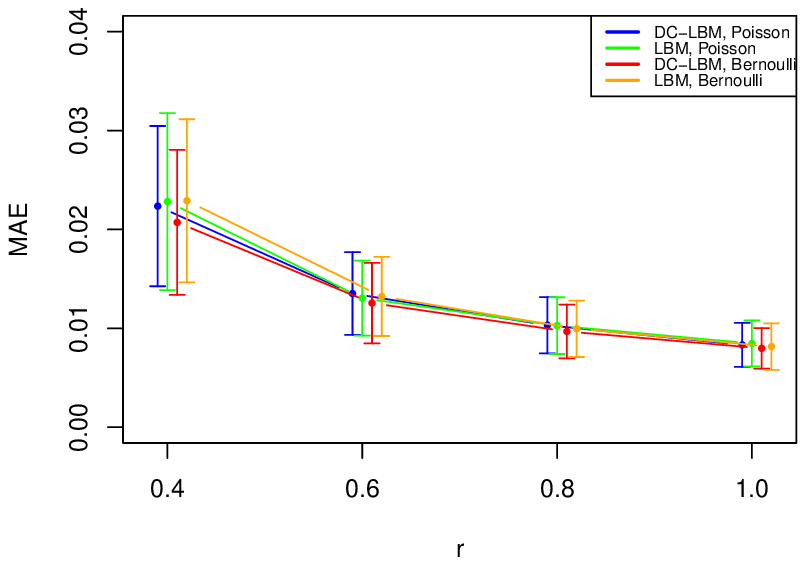}
 \end{center}
 \vspace{-0.5cm}
 \caption{MAEs for estimates of $\mu$ by the variational EM algorithm. $r$: the graph density factor in \eqref{mu_simulation}. } 
 \label{fig:mu_MAE}
\end{figure}

Finally, we report the performance of the estimations of the block-wise mean parameters $\mu$ using the variational EM algorithm under the four scenarios, namely, DC-LBM and LBM with the Poisson distribution, and DC-LBM and LBM with the Bernoulli distribution. The estimates $\hat{\mu}$ in Algorithm \ref{alg:EM} are for the canonical parameters $\mu^{*}$. We therefore transform $\mu$ in our simulation setups to be the canonical parameters. Another benefit of this transformation is that the canonical parameters $\mu^{*}$ have the comparable scale under different density levels $r$. Moreover, note that the cluster labels are subject to permutations. Therefore, we choose the best match of the estimated row and column cluster labels with the true labels among all possible permutations and rearrange ${\hat{\mu}_{kl}}$ accordingly. We report the mean absolute deviations (MAEs) in Figure \ref{fig:mu_MAE}, where each point represents the MAE over 12 parameters and 200 replicates. The performance exhibits a consistent pattern across the four scenarios; that is, all MAEs decrease as the graph density grows. This demonstrates an improvement in the performance of the estimations of the canonical parameters with increasing graph density.

\section{Application to  MovieLens data}\label{sec:data}
In this section we apply the proposed method to the well-known MovieLens data set 
\citep{harper2015movielens}. The data was collected by a research team at the University of Minnesota through the MovieLens website (https://movielens.org/) during a seven-month period from September 19th, 1997 to April 22nd, 1998. The data set contains 100,000 ratings from 943 users on 1682 movies. Demographic information for the users were also available in the data set, which is not used in the present analysis. The goal of the analysis is to simultaneously identify the group structure of the users and of the movies with the expectation that it can reveal patterns of consumer behavior such as which group of users like to watch which types of movies. 

We compare our method with PL that was applied to the same data set \citep{flynn2020profile}. As in \cite{flynn2020profile}, we constructed a 943-by-1682 binary matrix $A$ where $A_{ij}=1$ if user $i$ has rated movie $j$ and $A_{ij}=0$ otherwise. We chose the number of user clusters $K=3$ and the number of movie clusters $L=4$ and ran PL with the Bernoulli link function and 250 random initial values, as described in \cite{flynn2020profile}. The cluster numbers were chosen by \cite{flynn2020profile} based on the visualization of likelihoods in scree plots. We used the same cluster numbers, $K=3$ and $L=4$, when fitting the DC-LBM to the data for a fair comparison. 
\begin{figure}[!ht]
	\twoImages{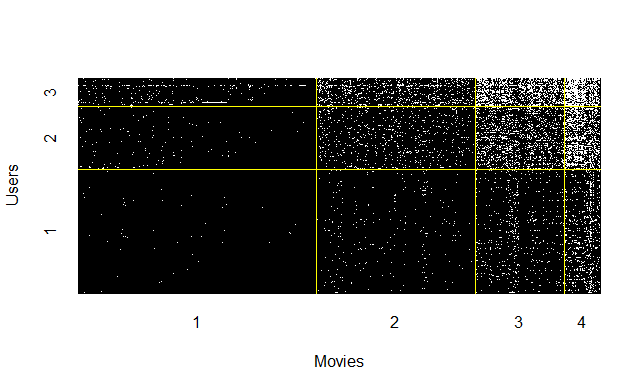}{8cm}{(a) PL}{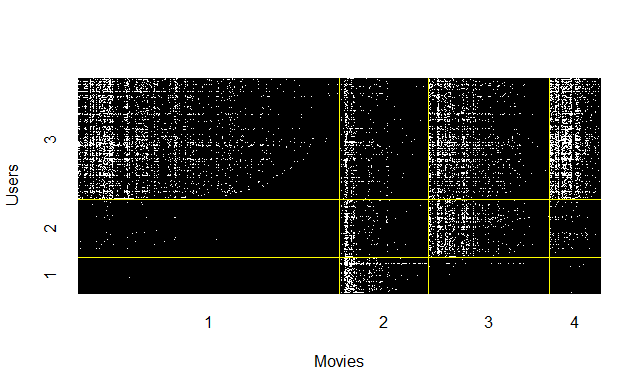}{8cm}{(b) DL-LBM}
	\caption{Heatmaps on the MovieLens matrix $A$ with rows and columns rearranged by biclustering results. White pixels: $A_{ij}=1$. Black pixels: $A_{ij}=0$. Yellow lines: boundaries of estimated clusters.}
	\label{fig:heatmap}
\end{figure}

Figure \ref{fig:heatmap} presents the heatmaps of the data matrix with rows and columns rearranged based on the biclustering results from PL and DC-LBM, respectively. From the left panel, we can see that the label assignment by PL is largely dominated by marginal information on rows and columns, i.e., row and column degrees. The DC-LBM, by contrast, allows a higher of level of degree heterogeneity within cluster. The biclustering result by DC-LBM reveals certain patterns of consumer behavior---for example, 
users in cluster 1 almost only reviewed movies in cluster 2, and movies in clusters 1 were primarily reviewed by users in cluster 3.  

\begin{figure}[!ht]
	\threeImages{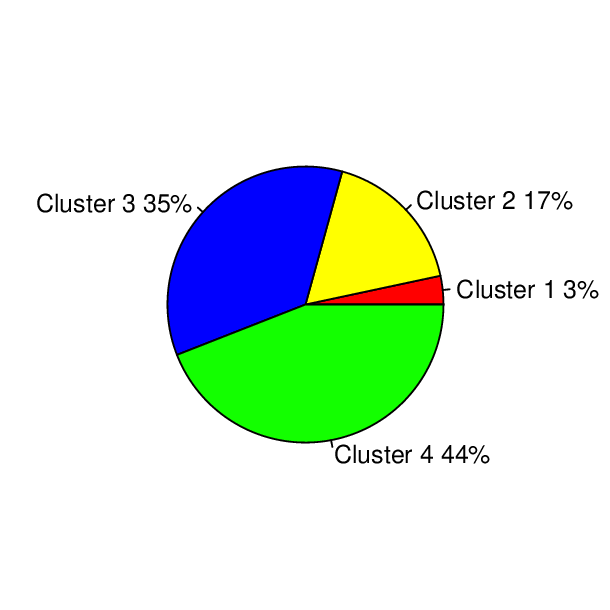}{5cm}{(a) User cluster 1 by PL }{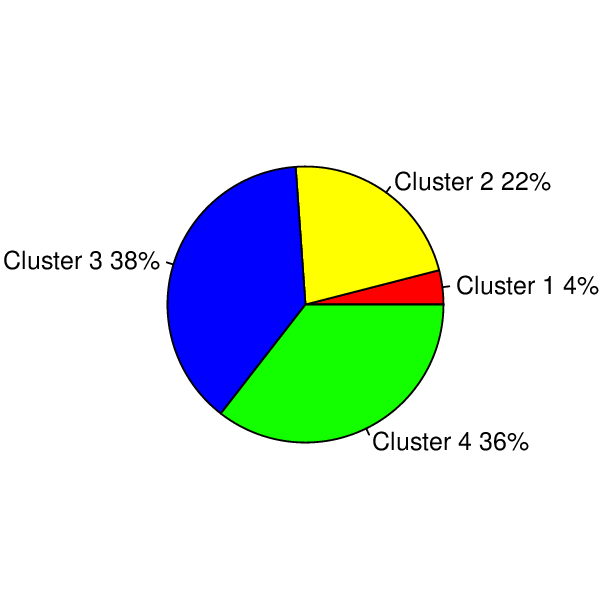}{5cm}{(b) User cluster 2 by PL}{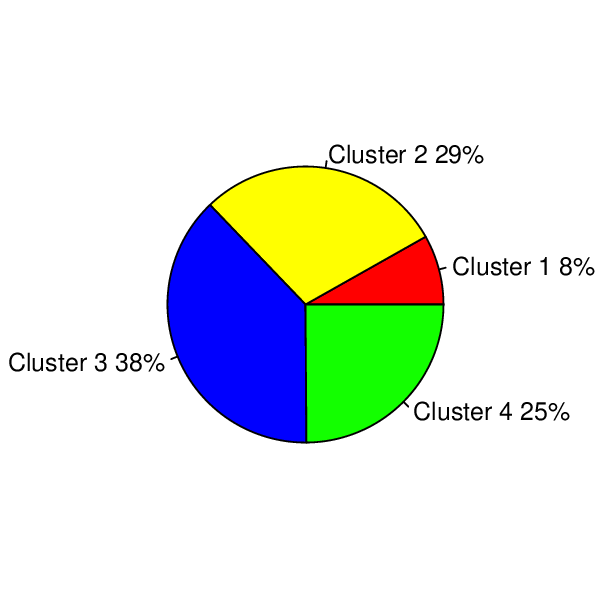}{5cm}{(c) User cluster 3 by PL}
	\threeImages{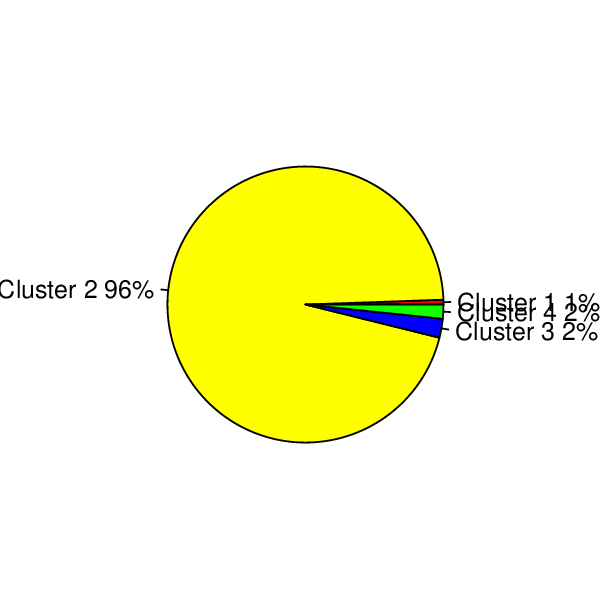}{5cm}{(d) User cluster 1 by DC-LBM }{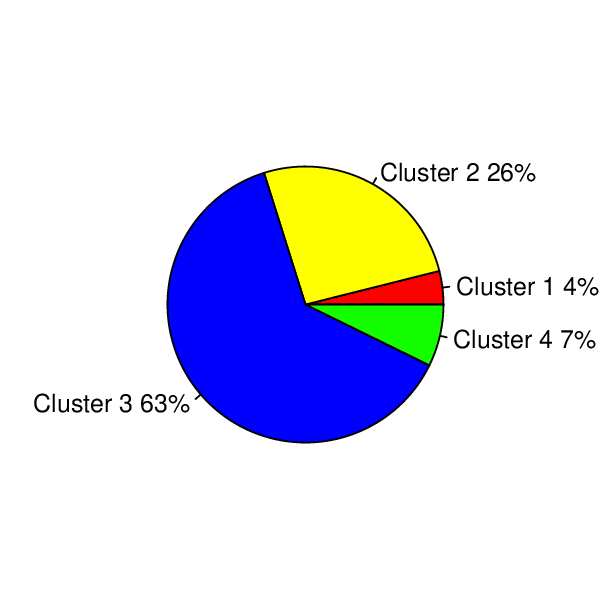}{5cm}{(e) User cluster 2 by DC-LBM}{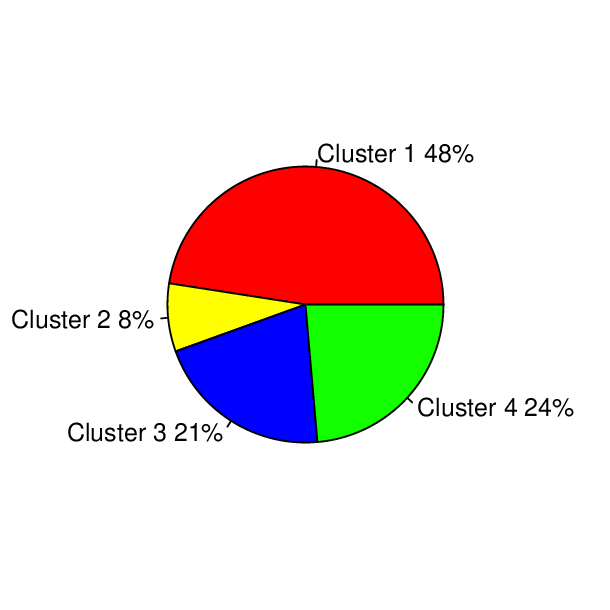}{5cm}{(f) User cluster 3 by DC-LBM}
	\caption{Pie charts showing the percentages of movies in each movie cluster that are rated by user clusters identified by PL and DC-LBM, respectively. }
	\label{fig:pie}
\end{figure}

The difference in user habits from different groups is better illustrated by the following pie charts in Figure \ref{fig:pie}. The pie charts show the percentages of movies in each movie cluster that are rated by user clusters identified by PL and DC-LBM, respectively. For example, panel (a) indicates that movies in cluster 4  by PL occupy 44\% of total movies rated by user cluster 1. Pie charts for DC-LBM show a much more heterogeneous pattern on percentages across the three user clusters than the pie charts for PL. Specifically, user cluster 1 mainly rated movie cluster 2; user cluster 2 mainly rated movie cluster 3;  user cluster 3 mainly rated movie clusters 1 and 4.

\begin{figure}[!ht]
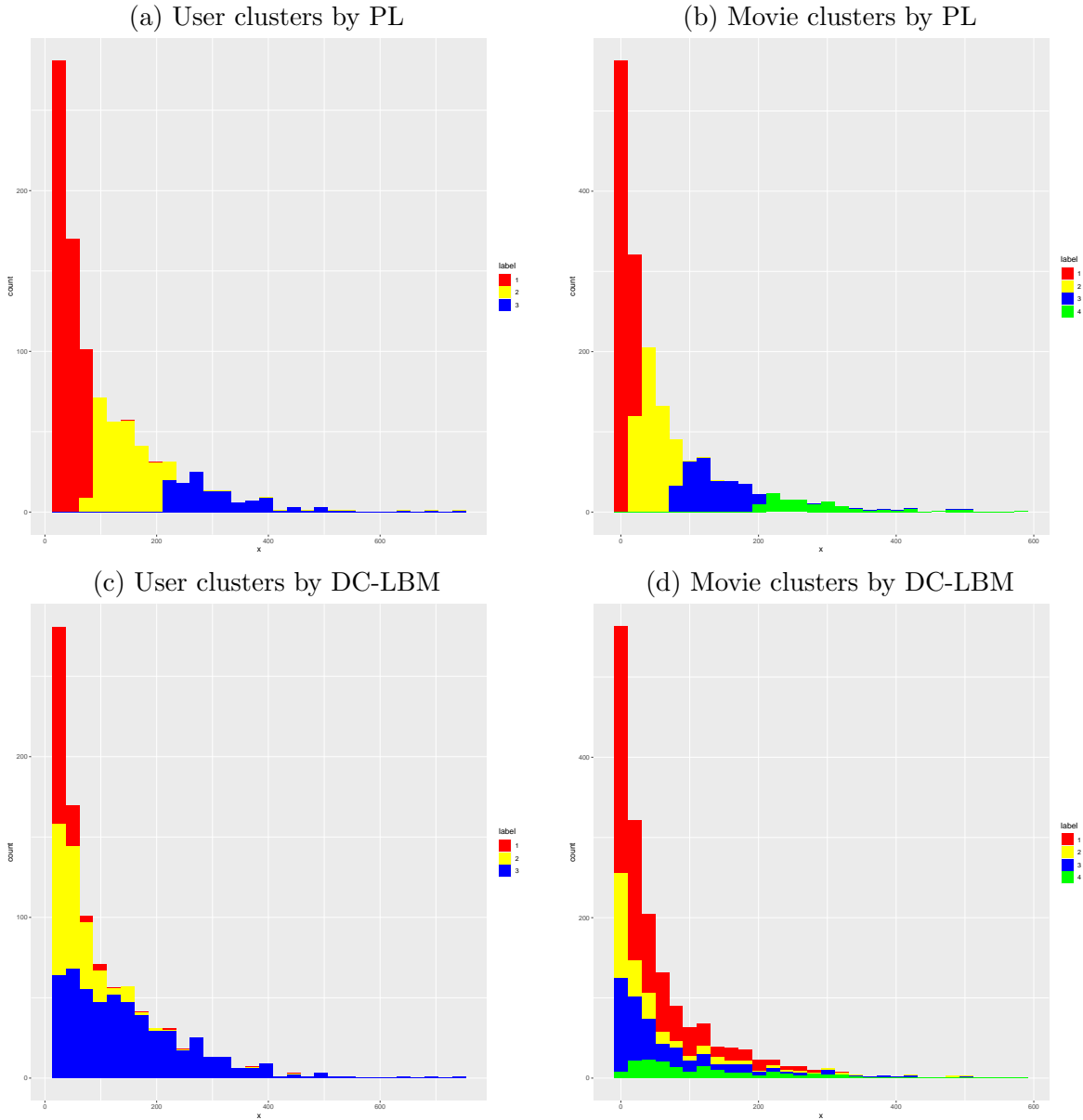

	\twoImages{new_row_profile.eps}{7.5cm}{(a) User clusters by PL }{new_col_profile.eps}{7.5cm}{(b) Movie clusters by PL}
		\twoImages{new_row_variational.eps}{7.5cm}{(c) User clusters by DC-LBM }{new_col_variational.eps}{7.5cm}{(d) Movie clusters by DC-LBM}
	\caption{Histograms of degrees in user and movie clusters identified by PL and DC-LBM, respectively. }
	\label{fig:hist}
\end{figure}

Furthermore, we compare the frequencies of degrees in user and movie clusters identified by PL and DC-LBM in Figure \ref{fig:hist}. As in the histograms, the degrees are much more homogeneous within a user or a movie cluster found by PL than in the clusters found by DC-LBM, which is in line with the observation from Figure \ref{fig:heatmap}. 

Finally, we study whether the estimated movie clusters are associated with the true movie categories provided in the MovieLens data set. The 1682 movies in the data set were labeled with 19 categories such as ``Action'' or ``Romance'' and many of the movies belong to multiple categories. A direct comparison between the estimated clusters and the true categories is difficult due to the relatively large number of movie categories and the overlaps. We instead construct contingency tables with row categories being the estimated movie clusters and column categories being the ground truth and evaluate how much the table deviates from an independence model. Specifically, we filtered the data to only include movies belonging to a single category,  resulting 833 movies, and constructed contingency tables for PL and DC-LBM, respectively. We then ran the chi-squared test of independence on the two tables. The p-values for the tables constructed from clusters estimated by PL and DC-LBM are $0.0415$ and $2.66 \times 10^{-7}$, respectively, which suggests that the movie clusters estimated by DC-LBM have a stronger association with the true movie categories.

\section{Conclusion}

In this paper, we proposed a degree-corrected latent block model (DC-LBM) for biclustering in bipartite networks. By introducing additional parameters to characterize row and column degrees, we achieved significant improvements in biclustering results compared to the classical LBM on both simulated and real-world data sets. We demonstrated that under the Poisson assumption, the maximizer of the variational approximation corresponds exactly to the row and column degrees. Furthermore, we established the label consistency and the convergence rate of the variational estimator under the DC-LBM with the Bernoulli and Poisson distributions, allowing for the expected graph density to approach zero as the average expected degrees of rows and columns go to infinity. For networks with weighted edges of more general types, label consistency is expected if the weights are nonnegative and bounded, although the assumption on graph densities may change depending on the variance of the weights. In the more general case, new concentration inequalities need to be developed (to replace \eqref{concentration_poisson} and the Bernstein inequality) to establish label consistency.

The proposed method can also be applied to clustering in directed networks, where row and column cluster labels are assumed to be distinct, capturing different behaviors of nodes as link senders and receivers. If the row and column cluster labels need to be identical, an additional step after the E step can be incorporated to enforce label probability assignment identity between rows and columns, similar to the split likelihood method \citep{wang2021efficient}. 

There are several directions for future work. One interesting area is the selection of the number of clusters in bipartite networks. In recent years, significant progress has been made  on the selection of communities in networks \citep{saldana2017many,wang2017likelihood,hu2019corrected,ma2021determining,le2022estimating,watanabe2021goodness}.  It is interesting to explore how to adapt these methods to DC-LBM. Additionally, we would like to explore the generalization of the DC-LBM to other clustering problems within the context of bipartite networks, such as estimating mixed memberships \citep{Airoldi2008,jin2017estimating,zhang2020overlapping} and incorporating node features \citep{zhang2016community,zhao2019logistic}. Furthermore, the theory in the paper guarantees that the global optimizer of \eqref{criterion} is consistent with the true cluster labels. However, the consistency of the solution produced by the proposed variational EM algorithm remains an open problem. Recent rigorous studies have delved into the consistency of EM algorithms \citep{Yu_EM} and the K-means algorithm \citep{lu2016statistical} within the context of classical cluster analysis. In the domain of network community detection, \cite{amini2013pseudo} established the consistency of the EM algorithm for a pseudo-likelihood in the Stochastic Block Model (SBM) with two communities. This method of proof has subsequently been adapted in \cite{zhao2019logistic,wang2021efficient,wang2021fast}. We plan to explore how to prove the consistency of EM algorithms to networks with more than two communities and with degree parameters in future work.

\appendix
\section{Proofs}
\label{app:theorem}
We give technical proofs in this section. 
\subsection{Proof of Proposition \ref{thm:stationary}}
	Up to a constant independent of $\Phi$,
	\begin{align*} 
	 & J(q_1,q_2,\Phi) \\
 = &   \sum_{{z}\in \Omega_{{z}}}\sum_{{w}\in \Omega_{{w}}} q_1({z}) q_2({w}) \left ( -\sum_{i=1}^m \sum_{j=1}^n  \theta_i \lambda_j  \left ( \sum_{k=1}^K \sum_{l=1}^L 1(z_i=k) 1(w_j=l) \mu_{k l} \right ) \right.   \\
	&  + \sum_{i=1}^m \sum_{j=1}^n  A_{ij} \left (\sum_{k=1}^K  \sum_{l=1}^L 1(z_i=k) 1(w_j=l) \log  \mu_{k l}  \right ) +\sum_{i=1}^m \sum_{j=1}^n  A_{ij} (\log  \theta_i+\log \lambda_j)\\
	&\left .+\sum_{i=1}^m  \sum_{k=1}^K 1(z_i=k)\log  \pi_{k} +\sum_{j=1}^n  \sum_{l=1}^L 1(w_j=l) \log \rho_{l} \right ) \\
	= &  -\sum_{i=1}^m \sum_{j=1}^n  \theta_i \lambda_j  \left (\sum_{k=1}^K \sum_{l=1}^L \mathbb{P}_{q_1}(z_i=k)\mathbb{P}_{q_2}(w_j=l) \mu_{k l} \right )+ \sum_{i=1}^m \sum_{j=1}^n  A_{ij} (\log  \theta_i+\log \lambda_j) \\
	& + \sum_{i=1}^m \sum_{j=1}^n  A_{ij} \left (\sum_{k=1}^K \sum_{l=1}^L \mathbb{P}_{q_1}(z_i=k)\mathbb{P}_{q_2}(w_j=l) \log  \mu_{kl}  \right )  \\
	& +  \sum_{i=1}^m  \left ( \sum_{k=1}^K \mathbb{P}_{q_1}(z_i=k) \log \pi_k \right )  + \sum_{j=1}^n \left ( \sum_{l=1}^L \mathbb{P}_{q_2}(w_j=l)   \log \rho_l \right ).
	\end{align*}
	The maximization of ${\pi}$ and ${\rho}$ is trivial. 
	
	We first prove	 $(\hat{{\theta}}, \hat{{\lambda}},\hat{\mu})$ defined in the proposition is a stationary point for any $q_1,q_2$.
	\begin{align*}
	& \left . \frac{\partial J}{\partial \theta_i} \right |_{\hat{{\theta}},\hat{{\lambda}},\hat{\mu}} \\
 = & - \sum_{j=1}^n   \hat{\lambda}_j  \left (\sum_{k=1}^K \sum_{l=1}^L \mathbb{P}_{q_1}(z_i=k)\mathbb{P}_{q_2}(w_j=l) \hat{\mu}_{k l} \right )+ \frac{\sum_{j=1}^n  A_{ij}}{\hat{\theta}_i} \\
	= & -\sum_{k=1}^K \sum_{l=1}^L \mathbb{P}_{q_1}(z_i=k) \left (\sum_{j=1}^n  d_{j}^c \mathbb{P}_{q_2}(w_j=l) \right )  \frac{\sum_{i'=1}^m \sum_{j'=1}^n  A_{i'j'}  \mathbb{P}_{q_1}(z_{i'}=k) \mathbb{P}_{q_2}(w_{j'}=l)}{\sum_{i'=1}^m \sum_{j'=1}^n d_{i'}^r d_{j'}^c \mathbb{P}_{q_1}(z_{i'}=k) \mathbb{P}_{q_2}(w_{j'}=l) } +1\\
	= &  -\sum_{k=1}^K \sum_{l=1}^L  \mathbb{P}_{q_1}(z_i=k) \frac{\sum_{i'=1}^m \sum_{j'=1}^n  A_{i'j'}  \mathbb{P}_{q_1}(z_{i'}=k) \mathbb{P}_{q_2}(w_{j'}=l)}{\sum_{i'=1}^m d_{i'}^r \mathbb{P}_{q_1}(z_{i'}=k)  } +1\\
	= & -\sum_{k=1}^K \mathbb{P}_{q_1}(z_i=k) \frac{\sum_{i'=1}^m \sum_{j'=1}^n  A_{i'j'}  \mathbb{P}_{q_1}(z_{i'}=k) \sum_{l=1}^L  \mathbb{P}_{q_2}(w_{j'}=l)}{\sum_{i'=1}^m d_{i'}^r \mathbb{P}_{q_1}(z_{i'}=k)  }+1 \\
	= & - \sum_{k=1}^K \mathbb{P}_{q_1}(z_i=k) \frac{\sum_{i'=1}^m (\sum_{j'=1}^n  A_{i'j'})  \mathbb{P}_{q_1}(z_{i'}=k) }{\sum_{i'=1}^m d_{i'}^r \mathbb{P}_{q_1}(z_{i'}=k)  } +1\\
	= & -\sum_{k=1}^K \mathbb{P}_{q_1}(z_i=k) + 1 =0, \,\, i=1,...,m.
	\end{align*}
	Similarly,
	\begin{align*}
	\left . \frac{\partial J}{\partial \lambda_j} \right |_{\hat{{\theta}},\hat{{\lambda}},\hat{\mu}} =0, \,\, j=1,...,n.
	\end{align*}
	And it is easy to check that
	\begin{align*}
	\left . \frac{\partial J}{\partial \mu_{kl}} \right |_{\hat{{\theta}},\hat{{\lambda}},\hat{\mu}} =0, \,\, k=1,...,K,\,l=1,...,L. 
	\end{align*}
	Let $\alpha_i=\log \theta_i, \beta_j=\log \lambda_j$ and $\gamma_{kl}=\log \mu_{kl}$. Then as a function of ${\alpha},{\beta}$ and $\gamma$, $J(q_1,q_2,\Phi)$ has the form (omitting the last two terms that depend on only ${\pi}$ and ${\rho}$)
	\begin{align}
	J(q_1,q_2,\Phi) = &  -\sum_{i=1}^m \sum_{j=1}^n  \sum_{k=1}^K \sum_{l=1}^L \mathbb{P}_{q_1}(z_i=k)\mathbb{P}_{q_2}(w_j=l) \exp( \alpha_i+ \beta_j+\gamma_{k l})  \nonumber\\
	&+ \sum_{i=1}^m \sum_{j=1}^n  A_{ij} (\alpha_i+\beta_j) + \sum_{i=1}^m \sum_{j=1}^n  A_{ij} \left (\sum_{k=1}^K \sum_{l=1}^L \mathbb{P}_{q_1}(z_i=k)\mathbb{P}_{q_2}(w_j=l)\gamma_{kl}  \right ).\label{temp10}
	\end{align}
	It is easy to see  $J(q_1,q_2,\Phi)$ is concave by noticing that $ \exp(\alpha_i+ \beta_j+\gamma_{k l}) $ is convex by definition and  the last two terms in $J(q_1,q_2,\Phi)$ are linear.

	By the chain rule, $(\hat{{\alpha}},\hat{{\beta}},\hat \gamma)$ with $\hat{\alpha}_i=\log \hat{\theta}_i, \hat{\beta}_j=\log \hat{\lambda}_j$ and $\hat{\gamma}_{kl}=\log \hat{\mu}_{kl}$ is also a stationary point of $J(q_1,q_2,\Phi)$, and therefore a global maximizer by concavity. This implies $(\hat{{\theta}},\hat{{\lambda}},\hat{\mu})$ is a global maximizer. 

	Now we work on the uniqueness of the maximizer. First, we need a lemma.
 \begin{lemma}\label{lemma2}
	For any two maximizers $(\hat{{\alpha}},\hat{{\beta}},\hat\gamma)$, $(\tilde{{\alpha}},\tilde{{\beta}},\tilde\gamma)$ of \eqref{temp10}, we have $\hat\alpha_i+\hat\beta_j+\hat\gamma_{kl}=\tilde\alpha_i+\tilde\beta_j+\tilde\gamma_{kl}$ when $\mathbb{P}_{q_1}(z_i=k)\mathbb{P}_{q_2}(w_j=l)\ne0$.
\end{lemma}
\begin{proof}
	By concavity of $J$, any vector in the segment connecting  two maximizers $(\hat{{\alpha}},\hat{{\beta}},\hat\gamma)$ and $(\tilde{{\alpha}},\tilde{{\beta}},\tilde\gamma)$ is also a maximizer. That is, 
	$(\hat{{\alpha}},\hat{{\beta}},\hat\gamma)+t(\tilde{{\alpha}}-\hat{{\alpha}},\tilde{{\beta}}-\hat{{\beta}},\tilde\gamma-\hat\gamma)$ is a maximizer for $0\leq t\leq 1$. This implies $J$ is a constant on the segment. Define $\check J(t)=J(q_1,q_2,\hat\Phi+t(\tilde \Phi-\hat\Phi))$. Therefore, $\check J$ is constant over $[0,1]$ and $\check J''|_{t=0}=0$. In fact, we have
	\begin{align*}
	& \check J''|_{t=0} = \\
 & -\sum_{i=1}^m \sum_{j=1}^n  \sum_{k=1}^K \sum_{l=1}^L \mathbb{P}_{q_1}(z_i=k)\mathbb{P}_{q_2}(w_j=l) (\hat\alpha_i+ \hat\beta_j+\hat\gamma_{k l}-\tilde\alpha_i-\tilde\beta_j-\tilde\gamma_{kl})^2\exp( \hat\alpha_i+ \hat\beta_j+\hat\gamma_{k l}) 
	\end{align*}
	$\check J''|_{t=0}=0$ implies $\mathbb{P}_{q_1}(z_i=k)\mathbb{P}_{q_2}(w_j=l) (\hat\alpha_i+ \hat\beta_j+\hat\gamma_{k l}-\tilde\alpha_i-\tilde\beta_j-\tilde\gamma_{kl})^2=0$ for all $i,j,k,l$, which leads to the conclusion. 
\end{proof}
	
 For a maximizer of \eqref{temp10}, say, $(\hat{{\alpha}},\hat{{\beta}},\hat\gamma)$, it is straightforward to check $(\hat{{\alpha}}+c_1{1}_m,\hat{{\beta}}+c_2{1}_n,\hat\gamma+{1}_K{1}_L^{T})$ is also a maximizer, for any constant $c_1$ and $c_2$. Here ${1}_{m}$ is a $m$-dimensional vector with all entries equal to 1. The argument below implies all maximizers are of the form $(\hat{{\alpha}}+c_1{1}_m,\hat{{\beta}}+c_2{1}_n,\hat\gamma+{1}_K{1}_L^{T})$ if $\mathbb{P}_{q_1}(z_i=k)\ne0$ for all $i$ and $k$, and $\mathbb{P}_{q_2}(w_j=l)\ne0$ for all $j$ and $l$. In terms of original parametrization, all maximizers are of the form $(e^{c_1}\hat{{\theta}},e^{c_2}\hat{{\lambda}},e^{-c_1-c_2}\hat\mu)$.
	
	Consider two maximizers $(\hat{{\alpha}},\hat{{\beta}},\hat\gamma)$, $(\tilde{{\alpha}},\tilde{{\beta}},\tilde\gamma)$ of \eqref{temp10}. For $i$ and $i'$, if there is a $k$ such that $\mathbb{P}_{q_1}(z_i=k)\mathbb{P}_{q_1}(z_{i'}=k)\ne0$, we show $\hat\alpha_i-\tilde\alpha_i=\hat\alpha_{i'}-\tilde\alpha_{i'}$. First find a $j$ and an $l$ with $\mathbb{P}_{q_2}(w_{j}=l)\ne0$. By Lemma \ref{lemma2} below, we have 
	$\hat\alpha_i+\hat\beta_j+\hat\gamma_{kl}=\tilde\alpha_i+\tilde\beta_j+\tilde\gamma_{kl}$ and $\hat\alpha_{i'}+\hat\beta_j+\hat\gamma_{kl}=\tilde\alpha_{i'}+\tilde\beta_j+\tilde\gamma_{kl}$, which implies $\hat\alpha_i-\tilde\alpha_i=\hat\alpha_{i'}-\tilde\alpha_{i'}$. We can make a similar conclusion for $\hat{{\beta}}$ and $\tilde{{\beta}}$. As a consequence, if  $\mathbb{P}_{q_1}(z_i=k)\ne0$ for all $i$ and $k$, $\hat\alpha_i-\tilde\alpha_i$ is constant for all $i$; if  $\mathbb{P}_{q_2}(w_j=l)\ne0$ for all $j$ and $l$, $\hat\beta_j-\tilde\beta_j$ is constant for all $j$. If there is a partition of the index set $\{1,...,m\}=\coprod_{s=1}^S\mathcal{C}_s$ such that $(i)$ $\mathbb{P}_{q_1}(z_i=k)\mathbb{P}_{q_1}(z_i'=k)=0$ for all $k$ whenever $i\in\mathcal{C}_s$ and $i'\in\mathcal{C}_{s'}$ with $s\ne s'$; $(ii)$ for $i,i'\in \mathcal{C}_s$, there is a $k$ such that $\mathbb{P}_{q_1}(z_i=k)\mathbb{P}_{q_1}(z_i'=k)\ne0$;  then we have $\hat\alpha_i-\tilde\alpha_i=d_s$ when $i\in\mathcal{C}_s$. That is, $\hat\alpha_i-\tilde\alpha_i$ can take $S$ different values based on the membership of $i$ with respect to the partition. We can make a similar conclusion for $\hat{{\beta}}$ and $\tilde{{\beta}}$. 

\subsection{Proof of Proposition \ref{thm:E}}
	We only prove the result for $q_1$ and the other part is similar. Recall that 
	\begin{align*}
	g_1(z_i) = &  -\sum_{j=1}^n   \theta_i \lambda_j \left ( \sum_{l=1}^L \mathbb{P}_{q_2}(w_j=l) \mu_{z_i l} \right )+ \sum_{j=1}^n  A_{ij} \left ( \sum_{l=1}^L \mathbb{P}_{q_2}(w_j=l) \log  \mu_{z_i l}  \right )+ \log \pi_{z_i}, \\
    & \quad \quad \quad \quad \quad \quad \quad \quad \quad \quad \quad \quad \quad \quad \quad \quad \quad \quad \quad \quad \quad \quad \quad \quad  \quad \quad \quad \quad \quad \quad i=1,...,m.
	\end{align*}
	Up to a constant independent of $q_1$,
	\begin{align*}
	J(q_1,q_2,\Phi) = &  \sum_{{z}\in \Omega_{{z}}} \sum_{{w}\in \Omega_{{w}}} q_1({z}) q_2({w}) \log P(A,{z},{w}) - \sum_{{z}\in \Omega_{{z}}} q_1({z}) \log q_1({z}) \\
	= &  \sum_{{z}\in \Omega_{{z}}}\sum_{{w}\in \Omega_{{w}}} q_1({z}) q_2({w}) \left ( -\sum_{i=1}^m \sum_{j=1}^n  \theta_i \lambda_j  \left ( \sum_{l=1}^L 1(w_j=l) \mu_{z_i l} \right ) \right.  \\
	& \left. + \sum_{i=1}^m \sum_{j=1}^n  A_{ij} \left ( \sum_{l=1}^L 1(w_j=l) \log  \mu_{z_i l}  \right ) +\sum_{i=1}^m \log \pi_{z_i}\right )- \sum_{{z}\in \Omega_{{z}}} q_1({z}) \log q_1({z}) \\
	= &  \sum_{{z}\in \Omega_{{z}}} q_1({z})  \left ( -\sum_{i=1}^m \sum_{j=1}^n  \theta_i \lambda_j  \left ( \sum_{l=1}^L \mathbb{P}_{q_2}(w_j=l) \mu_{z_i l} \right ) \right.  \\
	& \left. + \sum_{i=1}^m \sum_{j=1}^n  A_{ij} \left ( \sum_{l=1}^L \mathbb{P}_{q_2}(w_j=l) \log  \mu_{z_i l}  \right ) +\sum_{i=1}^m \log \pi_{z_i}\right )- \sum_{{z}\in \Omega_{{z}}} q_1({z}) \log q_1({z}) \\
	= & \sum_{{z}\in \Omega_{{z}}} q_1({z}) \sum_{i=1}^m g_1(z_i)- \sum_{{z}\in \Omega_{{z}}} q_1({z}) \log q_1({z}) \\
	= & \sum_{{z}\in \Omega_{{z}}} q_1({z}) \log \left \{  \frac{ \prod_{i=1}^m e^{g_1(z_i)}}{q_1({z}) }\right \} \\
	\leq & \log\prod_{i=1}^m  \left (  \sum_{z_i=1 }^K e^{g_1(z_i)} \right ),
	\end{align*}
	where the last inequality is Jensen's inequality and equality holds if and only if
	\begin{align*}
	q_1({z}) = & \prod_{i=1}^m \frac{ e^{g_1(z_i)} }{ \sum_{k=1}^K  e^{g_1(k)} }.
	\end{align*} 

\subsection{Proof of Proposition \ref{thm:well_defined}}
	Let ${\theta}, {\lambda}$ and $\mu$ be a set of arbitrarily chosen parameters that gives (5). Then
	\begin{align*}
	\theta^*_i & =   \theta_i \frac{\frac{1}{n}\sum_{j=1}^n \lambda_j \mu_{k w^*_j}}{\sqrt{\frac{1}{mn} \sum_{i=1}^m\sum_{j=1}^n E[A_{ij}|z_i^*,w_j^*] }}, \,\, \textnormal{for } z^*_i=k, \\
	\lambda^*_j & =  \lambda_j \frac{\frac{1}{m}\sum_{i=1}^m  \theta_i \mu_{z^*_i l}}{\sqrt{\frac{1}{mn} \sum_{i=1}^m\sum_{j=1}^n E[A_{ij}|z_i^*,w_j^*] }}, \,\, \textnormal{for } w_j^*=l,\\
	\mu^*_{kl} &= \mu_{kl}\frac{\sum_{i=1}^m\sum_{j=1}^n E[A_{ij}|z_i^*,w_j^*]}{\left(\sum_{j=1}^n \lambda_j \mu_{k w^*_j}\right)\left( \sum_{i=1}^m  \theta_i \mu_{z^*_i l}\right)}.
	\end{align*}
	
	The proposition immediately follows by noticing  $\sum_{j=1}^n \lambda_j \mu_{k w^*_j}$ depends only on $k$ and $\sum_{i=1}^m  \theta_i \mu_{z^*_i l}$ depends only on $l$. 

\subsection{Proof of Proposition \ref{thm:iden_cond}}
	Let ${\theta}, {\lambda}$ and $\mu$ be a set of arbitrarily chosen parameters that gives (5).  According to Proposition 3,
	\begin{align*}
	\theta^*_i & =   \theta_i \alpha_k, \,\, \textnormal{for } z^*_i=k, \\
	\lambda^*_j & =  \lambda_j \beta_l, \,\, \textnormal{for } w_j^*=l, \\
	\mu^*_{kl} & = \frac{\mu_{kl}}{\alpha_k \beta_l},
	\end{align*}
	where 
	\begin{align*}
	\alpha_k & = \frac{\frac{1}{n}\sum_{j=1}^n \lambda_j \mu_{k w^*_j}}{\sqrt{\frac{1}{mn} \sum_{i=1}^m\sum_{j=1}^n E[A_{ij}|z_i^*,w_j^*] }}, \\
	\beta_l & =  \frac{\frac{1}{m}\sum_{i=1}^m  \theta_i \mu_{z^*_i l}}{\sqrt{\frac{1}{mn} \sum_{i=1}^m\sum_{j=1}^n E[A_{ij}|z_i^*,w_j^*] }}.
	\end{align*}
	
	Now we show that $\mu^*_{kl}=\mu^*_{k'l}$ for all $l$ if and only if $\mu_{kl}/{\mu_{k'l}}$ is constant for all $l$, which leads to the conclusion of this proposition. 
	On one hand, 
	\begin{align*}
	& \mu^*_{kl}=\mu^*_{k'l} \,\Rightarrow \,\frac{\mu_{kl}}{\alpha_k \beta_l} =  \frac{\mu_{k'l}}{\alpha_k' \beta_l}
	\,\Rightarrow \, \frac{\mu_{kl}}{\mu_{k'l}}= \frac{\alpha_k}{\alpha_{k'}} , \,\, \textnormal{for all $l$}.
	\end{align*}
	On the other hand, if $\mu_{kl}/{\mu_{k'l}}=a_{kk'}$ for all $l$, it is straightforward to check $\alpha_k/{\alpha_{k'}}=a_{kk'}$, which implies $\mu^*_{kl}=\mu^*_{k'l}$. 

\subsection{Proof of Theorem \ref{thm:uniform}}
We first define a few  quantities. Let 
\begin{align*}
J_1(q^{{z}},q^{{w}},\mu) & = \sum_{i=1}^m \sum_{j=1}^n   A_{ij} \left (\sum_{k=1}^K \sum_{l=1}^L q^{{z}}_{ik} q^{{w}}_{jl} \log  \mu_{kl}  \right ), \\
\bar{J}_1(q^{{z}},q^{{w}},\mu) & =\sum_{i=1}^m \sum_{j=1}^n   E [A_{ij} |{z}^*,{w}^*]  \left (\sum_{k=1}^K \sum_{l=1}^L q^{{z}}_{ik} q^{{w}}_{jl} \log  \mu_{kl}  \right ), \\
J_2(q^{{z}},q^{{w}},\mu) & = -\sum_{i=1}^m \sum_{j=1}^n  \hat{\theta}_i \hat{\lambda}_j  \left (\sum_{k=1}^K \sum_{l=1}^L q^{{z}}_{ik} q^{{w}}_{jl} \mu_{k l} \right ), \\
\bar{J}_2(q^{{z}},q^{{w}},\mu) & = -\sum_{i=1}^m \sum_{j=1}^n  \theta_i^* \lambda_j^*  \left (\sum_{k=1}^K \sum_{l=1}^L q^{{z}}_{ik} q^{{w}}_{jl} \mu_{k l} \right ), \\
J_3(q^{{z}},q^{{w}},{\pi},{\rho}) & = \sum_{i=1}^m  \left ( \sum_{k=1}^K  q^{{z}}_{ik} \log \pi_k \right )  + \sum_{j=1}^n \left ( \sum_{l=1}^L  q^{{w}}_{jl}  \log \rho_l \right ) \\
& \quad - \sum_{i=1}^m  \sum_{k=1}^K q^{{z}}_{ik} \log q^{{z}}_{ik} -\sum_{j=1}^n \sum_{l=1}^L q^{{w}}_{jl} \log q^{{w}}_{jl}.
\end{align*}
It is easy to check that 
\begin{align*}
    \hat{J}(q^{{z}},q^{{w}},\Phi) & = J_1(q^{{z}},q^{{w}},\mu) +J_2(q^{{z}},q^{{w}},\mu) +J_3(q^{{z}},q^{{w}},{\pi},{\rho}), \\
    \bar{J}(q^{{z}},q^{{w}},\mu) & =\bar{J}_1(q^{{z}},q^{{w}},\mu)+\bar{J}_2(q^{{z}},q^{{w}},\mu). 
\end{align*}
\begin{lemma}\label{thm:general_poisson}
	Let $\{X_{ij}\}$ be independent Poisson variables with mean\footnote{All constants, such as $C_1$ and $C_2$, are defined locally. This means that the constants in different lemmas can vary.} $E[X_{ij}] \leq r_{mn} C$. Then for all $\epsilon>0$,
	\begin{align*}
	\mathbb{P} \left ( \max_{0\leq u_i \leq 1, i=1,...,m,  0 \leq v_j \leq 1,j=1,...,n}  \left |   \sum_{i=1}^m \sum_{j=1}^n  (X_{ij}-E[X_{ij}])  u_i v_j  \right | \geq mn r_{mn}  \epsilon \right ) \\
    \leq  2^{m+n+1}  \exp \left ( -\frac{m n r_{mn} \epsilon^2}{4 \max(C,\epsilon)} \right ).
	\end{align*}
\end{lemma}

\begin{proof}
	First note that \begin{align*}
	    & \,\, \max_{0\leq u_i \leq 1, i=1,...,m,  0 \leq v_j \leq 1,j=1,...,n}\left |   \sum_{i=1}^m \sum_{j=1}^n  (X_{ij}-E[X_{ij}])  u_i v_j  \right | \\
	 = & \,\, \max_{u_i\in \{0, 1\},i=1,...,m, v_j \in \{0,1\},j=1,...,n} \left |   \sum_{i=1}^m \sum_{j=1}^n  (X_{ij}-E[X_{ij}])  u_i v_j  \right | .
		\end{align*}

	To prove this, let $f({u},{v})=\sum_{i=1}^m \sum_{j=1}^n  (X_{ij}-E[X_{ij}])  u_i v_j$. If the maximum occurs at $({u},{v})$ with $u_1\in(0,1)$, then consider ${u}'$ and ${u}''$ with $u'_1=0$, $u''_1=1$, and rest of entries identical to that of ${u}$. It is easy to check $f({u},{v})=(1-u_1)f({u'},{v})+u_1f({u''},{v})$. So the value of $f({u},{v})$ must be between $f({u'},{v})$ and $f({u''},{v})$. This implies that both $({u'},{v})$ and $({u''},{v})$ are also maximizers. We can consider $({u'},{v})$ instead of $({u},{v})$. If there are other entries of $({u'},{v})$ strictly between 0 and 1, we can continue this argument until we find a maximizer with all entries equal to 0 or 1.
	
	Then according to \cite{Canonne_Poisson},
	for $u_i\in \{0, 1\},i=1,...,m, v_j \in \{0,1\},j=1,...,n$,
	\begin{align}
	& \mathbb{P} \left (  \left |   \sum_{i=1}^m \sum_{j=1}^n  (X_{ij}-E[X_{ij}])  u_i v_j  \right | \geq mn r_{mn} \epsilon \right ) \nonumber \\
	\leq & \,\, 2 \exp \left ( -\frac{m^2 n^2 r_{mn}^2 \epsilon^2}{2(\sum_{ij}  E[X_{ij}]+mn r_{mn} \epsilon)} \right ) \leq 2 \exp \left ( -\frac{m n r_{mn} \epsilon^2}{4\max(C,\epsilon)} \right ). \label{concentration_poisson}
	\end{align}
It follows that 
	\begin{align*}
	& \mathbb{P} \left ( \max_{0\leq u_i \leq 1, i=1,...,m,  0 \leq v_j \leq 1,j=1,...,n}  \left |   \sum_{i=1}^m \sum_{j=1}^n  (X_{ij}-E[X_{ij}])  u_i v_j  \right | \geq  mn r_{mn} \epsilon \right )  \\
	= & \,\, \mathbb{P} \left (  \max_{u_i\in \{0, 1\},i=1,...,m, v_j \in \{0,1\},j=1,...,n}  \left |   \sum_{i=1}^m \sum_{j=1}^n  (X_{ij}-E[X_{ij}])  u_i v_j  \right | \geq mn r_{mn} \epsilon \right ) \\
	\leq & \,\, 2^{m+n} 2 \exp \left ( -\frac{m n r_{mn} \epsilon^2}{4 \max(C,\epsilon) } \right ).
	\end{align*}
\end{proof}

 \begin{lemma}\label{thm:J1}
For sufficiently small positive $\epsilon$,	\begin{align*}
	\mathbb{P} \left ( \left . \max_{q^{{z}}\in \mathcal{C}_{{z}},q^{{w}}\in \mathcal{C}_{{w}},\mu \in \mathcal{C}_\mu }\left | 	J_1(q^{{z}},q^{{w}},\mu)- \bar{J}_1(q^{{z}},q^{{w}},\mu)  \right | \geq mn r_{mn} \epsilon 
	\right | {z}^*,{w}^* \right ) \\
   \leq C_1 2^{m+n} \exp \left ( - C_2 m n r_{mn} \epsilon^2 \right ).
	\end{align*}
\end{lemma}

\begin{proof}
Note that 
\begin{align*}
 & \,\, \left | 	J_1(q^{{z}},q^{{w}},\mu)- \bar{J}_1(q^{{z}},q^{{w}},\mu) \right | \\
= & \,\, \left | \sum_{i=1}^m \sum_{j=1}^n  (A_{ij}-E[A_{ij}|{z}^*,{w}^*]) \left (\sum_{k=1}^K \sum_{l=1}^L q^{{z}}_{ik} q^{{w}}_{jl} \log   \mu_{kl}  \right ) \right | \\
= & \,\, \left |  \sum_{k=1}^K \sum_{l=1}^L \log  \mu_{kl}  \left ( \sum_{i=1}^m \sum_{j=1}^n  (A_{ij}-E[A_{ij}|{z}^*,{w}^*])  q^{{z}}_{ik} q^{{w}}_{jl} \right ) \right | \\
\leq & \,\, \sum_{k=1}^K \sum_{l=1}^L \left | \log \mu_{kl} \right | \left |  \sum_{i=1}^m \sum_{j=1}^n  (A_{ij}-E[A_{ij}|{z}^*,{w}^*])  q^{{z}}_{ik} q^{{w}}_{jl}  \right | \\
\leq & \,\, \max \{|\log \mu_{\textnormal{min}}|, |\log \mu_{\textnormal{max}}| \} \sum_{k=1}^K \sum_{l=1}^L \left |  \sum_{i=1}^m \sum_{j=1}^n  (A_{ij}-E[A_{ij}|{z}^*,{w}^*])  q^{{z}}_{ik} q^{{w}}_{jl} \right |.
\end{align*}
The lemma follows immediately from Lemma \ref{thm:general_poisson}.
\end{proof}
 
\begin{lemma}\label{thm:J2}
For sufficiently small positive $\epsilon>0$, 
\begin{align*}
 &	\mathbb{P} \left ( \left . \max_{q^{{z}}\in \mathcal{C}_{{z}},q^{{w}}\in \mathcal{C}_{{w}},\mu \in \mathcal{C}_\mu }\left | 	J_2(q^{{z}},q^{{w}},\mu)- \bar{J}_2(q^{{z}},q^{{w}},\mu)  \right | \geq mn r_{mn} \epsilon 
	\right | {z}^*,{w}^* \right ) \\
 &\leq C_{1} 2^m \exp \left ( -C_{2} mn r_{mn} \epsilon^2 \right ) +C_{3} 2^n \exp \left ( -C_{4} mn r_{mn} \epsilon^2 \right ). 
\end{align*}
\end{lemma}
\begin{proof}
By a similar argument as in Lemma \ref{thm:general_poisson}, 
\begin{align}
	\,\, & \mathbb{P} \left ( \left . \max_{  q^{{z}}_{ik} \in [0,1], i=1,...,m}  \left | \sum_{i=1}^m \sum_{j=1}^n ( A_{ij}-E[A_{ij}|z^*,w^*])q^{{z}}_{ik}   \right | \geq mn r_{mn} \epsilon  \right |{z}^*,{w}^* \right ) \nonumber \\
	= \,\, & \mathbb{P} \left ( \left . \max_{  q^{{z}}_{ik} \in \{0,1 \}, i=1,...,m}  \left | \sum_{i=1}^m \sum_{j=1}^n ( A_{ij}-E[A_{ij}|z^*,w^*]])q^{{z}}_{ik}   \right | \geq mn r_{mn}  \epsilon  \right |{z}^*,{w}^* \right ) \nonumber \\ 
	\leq \,\, &  2^{m+1} \exp  \left ( -\frac{m n r_{mn} \epsilon^2}{4 \max(C,\epsilon)} \right ). \label{temp1}
\end{align}

	Let $E^r_i= E[d_i^r | {z}^*,{w}^*] = \sum_{j=1}^n E [A_{ij} | {z}^*,{w}^*]$ and $E^c_j= E[d_j^c | {z}^*,{w}^*]= \sum_{i=1}^m E [A_{ij} | {z}^*,{w}^*]$. 
Note that
\begin{align}
	& \,\,\max_{q^{{z}} \in \mathcal{C}_{{z}},q^{{w}} \in \mathcal{C}_{{w}} }\left |  \sum_{i=1}^m \sum_{j=1}^n  \left (\frac{d_i^r}{n} \frac{E_j^c}{m} - \frac{E_i^r}{n}  \frac{E^c_j}{m} \right ) q^{{z}}_{ik} q^{{w}}_{jl} \right | \nonumber \\
	\leq &\,\, \max_{q^{{w}} \in \mathcal{C}_{{w}}} \left |\sum_{j=1}^n  \frac{E_j^c}{m} q^{{w}}_{jl} \right | \max_{q^{{z}}\in \mathcal{C}_{{z}}} \left | \sum_{i=1}^m \left (\frac{d_i^r}{n}-\frac{E_i^r}{n} \right)  q^{{z}}_{ik} \right | \nonumber \\
	\leq & \,\,n E_{\textnormal{max}} r_{mn}  \max_{q^{{z}}\in \mathcal{C}_{{z}}} \left | \sum_{i=1}^m \left (\frac{d_i^r}{n}-\frac{E_i^r}{n} \right)  q^{{z}}_{ik} \right | \nonumber \\
	\leq &\,\,   E_{\textnormal{max}} r_{mn}  \max_{q^{{z}}\in \mathcal{C}_{{z}}} \left | \sum_{i=1}^m \sum_{j=1}^n (A_{ij}-E[A_{ij}|{z}^*,{w}^*])  q^{{z}}_{ik} \right |. \label{temp2}
\end{align}
From \eqref{temp1} and \eqref{temp2}
\begin{align}
		 \mathbb{P} \left ( \left . \max_{q^{{z}} \in \mathcal{C}_{{z}},q^{{w}} \in \mathcal{C}_{{w}} }\left |  \sum_{i=1}^m \sum_{j=1}^n  \left (\frac{d_i^r}{n} \frac{E_j^c}{m} - \frac{E_i^r}{n}  \frac{E^c_j}{m} \right ) q^{{z}}_{ik} q^{{w}}_{jl} \right | \geq mn r_{mn}^2 \epsilon  \right |{z}^*,{w}^* \right ) \nonumber \\
  \leq C_5 2^m \exp \left ( -C_6 mn r_{mn} \epsilon^2 \right ). \label{temp3}
		\end{align}

Next, we analyze the term $\max_{q^{{z}} \in \mathcal{C}_{{z}},q^{{w}} \in \mathcal{C}_{{w}} }\left |  \sum_{i=1}^m \sum_{j=1}^n \left (\frac{d_i^r}{n} \frac{d_j^c}{m} - \frac{d_i^r}{n}  \frac{E^c_j}{m} \right  ) q^{{z}}_{ik} q^{{w}}_{jl} \right |$.
		
If $\max_{q^{{z}} \in \mathcal{C}_{{z}}} \left | \sum_{i=1}^m \left (\frac{d_i^r}{n}-\frac{E_i^r}{n} \right) q^{{z}}_{ik} \right |\leq m r_{mn}\epsilon$,
\begin{align}
 & \,\, \max_{q^{{z}} \in \mathcal{C}_{{z}},q^{{w}} \in \mathcal{C}_{{w}} }\left |  \sum_{i=1}^m \sum_{j=1}^n \left (\frac{d_i^r}{n} \frac{d_j^c}{m} - \frac{d_i^r}{n}  \frac{E^c_j}{m} \right  ) q^{{z}}_{ik} q^{{w}}_{jl} \right | \nonumber \\
	\leq & \,\, \max_{q^{{z}} \in \mathcal{C}_{{z}}} \left |\sum_{i=1}^m  \frac{d_i^r}{n} q^{{z}}_{ik} \right | \max_{q^{{w}}  \in \mathcal{C}_{{w}}} \left | \sum_{j=1}^n \left (\frac{d_j^c}{m}-\frac{E_j^c}{m} \right)  q^{{w}}_{jl} \right | \nonumber \\
	= & \,\, \max_{q^{{z}} \in \mathcal{C}_{{z}}} \left | \sum_{i=1}^m \left (\frac{d_i^r}{n}-\frac{E_i^r}{n} \right) q^{{z}}_{ik} + \sum_{i=1}^m \frac{E_i^r}{n} q^{{z}}_{ik}  \right |  \max_{q^{{w}} \in \mathcal{C}_{{w}}} \left | \sum_{j=1}^n \left (\frac{d_j^c}{m}-\frac{E_j^c}{m} \right)  q^{{w}}_{jl} \right | \nonumber \\
	\leq & \,\, (mr_{mn} \epsilon +m E_{\textnormal{max}} r_{mn})  \max_{q^{{w}}  \in \mathcal{C}_{{w}}} \left | \sum_{j=1}^n \left (\frac{d_j^c}{m}-\frac{E_j^c}{m} \right)  q^{{w}}_{jl} \right | \nonumber \\
	= & \,\, ( \epsilon +E_{\textnormal{max}} )r_{mn} \max_{q^{{w}}  \in \mathcal{C}_{{w}}}  \left | \sum_{i=1}^m \sum_{j=1}^n ( A_{ij}-E[A_{ij} |{z}^*,{w}^*])  q^{{w}}_{jl}\right |. \label{temp4}
	\end{align}		
Therefore, 
		\begin{align}
	& \,\, \mathbb{P} \left ( \left .  \max_{q^{{z}}\in \mathcal{C}_{{z}},q^{{w}}\in \mathcal{C}_{{w} }}\left |  \sum_{i=1}^m \sum_{j=1}^n  \left (\frac{d_i^r}{n} \frac{d_j^c}{m} - \frac{d_i^r}{n}  \frac{E^c_j}{m}  \right) q^{{z}}_{ik} q^{{w}}_{jl} \right | \geq mn r_{mn}^2 \epsilon  \right |{z}^*,{w}^* \right ) \nonumber \\ 
	\leq & \,\, \mathbb{P}  \left (  \left . \max_{q^{{z}}\in \mathcal{C}_{{z}}} \left |\sum_{i=1}^m \left ( \frac{d_i^r}{n} -\frac{E_i^r}{n} \right) q^{{z}}_{ik} \right |  \geq m r_{mn} \epsilon  \right |{z}^*,{w}^* \right )  \nonumber \\
	& \,\, +  \mathbb{P}  \left (  \left .( \epsilon +E_{\textnormal{max}} )r_{mn} \max_{q^{{w}}  \in \mathcal{C}_{{w}}}  \left | \sum_{i=1}^m \sum_{j=1}^n ( A_{ij}-E[A_{ij} |{z}^*,{w}^*])  q^{{w}}_{jl}\right |  \geq mn r_{mn}^2 \epsilon  \right |{z}^*,{w}^* \right ) \nonumber \\
    \leq & \,\, C_7 2^m \exp \left ( -C_8 mn r_{mn} \epsilon^2 \right )+C_9 2^n \exp \left ( -C_{10} mn r_{mn} \epsilon^2 \right ).
     \label{temp5}
	\end{align}
To sum up, from \eqref{temp3} and \eqref{temp5}, for sufficiently small positive $\epsilon$,
\begin{align*}
	& \mathbb{P} \left ( \left .  \max_{q^{{z}}\in \mathcal{C}_{{z}},q^{{w}}\in \mathcal{C}_{{w} }}\left |  \sum_{i=1}^m \sum_{j=1}^n  \left (\frac{d_i^r}{n} \frac{d_j^c}{m}-\frac{E^r_i}{n} \frac{E^c_j}{m} \right)  q^{{z}}_{ik} q^{{w}}_{jl} \right | \geq mn r_{mn}^2 \epsilon  \right |{z}^*,{w}^* \right ) \\
 & \leq C_5 2^m \exp \left ( -C_6 mn r_{mn} \epsilon^2 \right ) +C_7 2^m \exp \left ( -C_8 mn r_{mn} \epsilon^2 \right )+C_9 2^n \exp \left ( -C_{10} mn r_{mn} \epsilon^2 \right ).
\end{align*}
Next, 
\begin{align*}
	& \,\, \mathbb{P} \left ( \left . \left |D-E[D |{z}^*,{w}^*] \right| \geq r_{mn} \epsilon \right| {z}^*,{w}^* \right) \\
	= & \,\, \mathbb{P} \left (\left. \left | \frac{1}{mn}  \sum_{ij} (A_{ij}-E[ A_{ij}|{z}^*,{w}^*]) \right| \geq r_{mn} \epsilon \right | {z}^*,{w}^* \right ) \\
	\leq & \,\,  C_{11} \exp \left ( -C_{12} mn r_{mn} \epsilon^2 \right ).
\end{align*}
Recall that
\begin{align*}
    \hat{\theta}_i \hat{\lambda}_j = \frac{d_i^r d_j^c}{mnD} \quad \textnormal{and} \quad \theta_i^*\lambda_j^* = \frac{E_i^r E_j^c}{mnE[D |{z}^*,{w}^*]}. 
\end{align*}
Then by a similar argument for convergence of ratio of two random variables, 
\begin{align*}
	&	\mathbb{P} \left ( \left .  \max_{q^{{z}}\in \mathcal{C}_{{z}},q^{{w}}\in \mathcal{C}_{{w} }}\left |  \sum_{i=1}^m \sum_{j=1}^n  (\hat{\theta}_i\hat{\lambda}_j - \theta_i^* \lambda_j^*) q^{{z}}_{ik} q^{{w}}_{jl} \right |  \geq mn r_{mn} \epsilon  \right |{z}^*,{w}^* \right ) \\
  & \leq C_{13} 2^m \exp \left ( -C_{14} mn r_{mn} \epsilon^2 \right ) +C_{15} 2^n \exp \left ( -C_{16} mn r_{mn} \epsilon^2 \right ) 
\end{align*}

	Finally, the lemma holds since
	\begin{align*}
	& \,\, \left | 	J_2(q^{{z}},q^{{w}},\mu)- E [J_2(q^{{z}},q^{{w}},\mu) |{z}^*,{w}^*] \right | \\
	= & \,\, \left | \sum_{i=1}^m \sum_{j=1}^n  (\hat{\theta}_i \hat{\lambda}_j-\theta_i^* \lambda_j^* ) \left (\sum_{k=1}^K \sum_{l=1}^L q^{{z}}_{ik} q^{{w}}_{jl}   \mu_{kl}  \right ) \right | \\
	\leq & \,\,  \sum_{k=1}^K \sum_{l=1}^L   \mu_{kl}  \left |  \sum_{i=1}^m \sum_{j=1}^n  (\hat{\theta}_i \hat{\lambda}_j-\theta_i^* \lambda_j^* )  q^{{z}}_{ik} q^{{w}}_{jl}  \right | \\
	\leq & \,\,   \mu_{\textnormal{max}} \sum_{k=1}^K \sum_{l=1}^L \left |  \sum_{i=1}^m \sum_{j=1}^n  (\hat{\theta}_i \hat{\lambda}_j-\theta_i^* \lambda_j^*)  q^{{z}}_{ik} q^{{w}}_{jl} \right |.
	\end{align*}
\end{proof}

Note that
\begin{align*}
& \left | \hat{J}(q^{{z}},q^{{w}},\Phi)- \bar{J}(q^{{z}},q^{{w}},\mu) \right | \ \\
\leq & \left | 	J_1(q^{{z}},q^{{w}},\mu)- \bar{J}_1(q^{{z}},q^{{w}},\mu) \right | +\left | 	J_2(q^{{z}},q^{{w}},\mu)- \bar{J}_2(q^{{z}},q^{{w}},\mu) \right | 
+ \left |J_3(q^{{z}},q^{{w}},{\pi},{\rho}) \right |,
\end{align*}
and
\begin{align*}
  \max_{q^{{z}},q^{{w}},{\pi} \in \mathcal{C}_{{\pi}}, {\rho} \in \mathcal{C}_{{\rho}}} \left | J_3(q^{{z}},q^{{w}},{\pi},{\rho}) \right | =O(m+n).
\end{align*}
Furthermore,
\begin{align}
& \mathbb{P} \left ( \left .  \max_{q^{{z}}\in \mathcal{C}_{{z}},q^{{w}}\in \mathcal{C}_{{w}}, {\pi}\in \mathcal{C}_{{\pi}}, {\rho}\in \mathcal{C}_{{\rho}},\mu \in \mathcal{C}_{\mu}  } \left |\hat{J}(q^{{z}} ,q^{{w}},\Phi)-\bar{J}(q^{{z}},q^{{w}},\mu)\right | \geq mn r_{mn} \epsilon
	\right | {z}^*,{w}^* \right ) \nonumber \\
	& \leq \mathbb{P} \left ( \left .  \max_{q^{{z}}\in \mathcal{C}_{{z}},q^{{w}}\in \mathcal{C}_{{w}}, {\pi}\in \mathcal{C}_{{\pi}}, {\rho}\in \mathcal{C}_{{\rho}},\mu \in \mathcal{C}_{\mu}  } \left |(\hat{J}_1+\hat{J}_2)(q^{{z}} ,q^{{w}},\mu)-(\bar{J}_1+\bar{J}_2)(q^{{z}},q^{{w}},\mu)\right | +C(m+n) \right. \right. \nonumber \\
& \quad  \quad  \quad \geq    mn r_{mn}  \epsilon \bigg | {z}^*,{w}^* \bigg ). \label{uniform}
\end{align}
For \eqref{uniform} converges to 0,  it is sufficient to assume $(mn r_{mn} \epsilon)/(m+n) \rightarrow \infty$ and $(mn r_{mn} \epsilon^2)/(m+n) \rightarrow \infty$ by Lemmas \ref{thm:J1} and \ref{thm:J2}. But the first condition is implied by the second one. Therefore, Theorem \ref{thm:uniform} is proven.

\subsection{Proof of Theorem \ref{thm:separability}}
Let
\begin{align*}
G( q^{{z}},q^{{w}},\mu)=\sum_{k=1}^K \sum_{l=1}^L \sum_{k'=1}^K \sum_{l'=1}^L \mathbb{R}_{ kk'}({1}^{{z}^*},q^{{z}}) \mathbb{R}_{ ll'}({1}^{{w}^*},q^{{w}})  \textnormal{KL}(\mu^*_{kl},\mu_{k'l'}), 
\end{align*}
where $\textnormal{KL}(\mu^*_{kl},\mu_{k'l'})= \mu^*_{kl} \log ( \mu^*_{kl} /\mu_{k'l'})-(\mu^*_{kl}-\mu_{k'l'})$.

The next proposition shows that $\bar{J}(q^{{z}},q^{{w}},\mu)$ is maximized at the true label assignments and true parameters, and gives a lower bound of the difference between $\bar{J}(q^{{z}},q^{{w}},\mu)$ and the maximum in the form of the confusion matrices. 
\begin{proposition}\label{thm:max}
	For any $q^{{z}}$, $q^{{w}}$, and $\mu$,
	\begin{align*}
	\bar{J}({1}^{{z}^*},{1}^{{w}^*},\mu^*)-\bar{J}(q^{{z}},q^{{w}},\mu)  \geq    mn r_{mn} \theta_{\textnormal{min}} \lambda_{\textnormal{min}} G( q^{{z}},q^{{w}},\mu) \geq 0.
	\end{align*}
 \end{proposition}

\begin{proof}
A straightforward calculation shows that
	\begin{align*}
	\bar{J}_1(q^{{z}},q^{{w}},\mu) & =\sum_{i=1}^m \sum_{j=1}^n   E [A_{ij} |{z}^*,{w}^*]  \left (\sum_{k'=1}^K \sum_{l'=1}^L q^{{z}}_{ik'} q^{{w}}_{jl'} \log  \mu_{k'l'}  \right ) \\
	& = \sum_{i=1}^m \sum_{j=1}^n   \theta_i^*\lambda_j^*  \mu_{z_i^*w_j^*}^* \left (\sum_{k'=1}^K \sum_{l'=1}^L q^{{z}}_{ik'} q^{{w}}_{jl'} \log  \mu_{k'l'}  \right ) \\
	& =  \sum_{i=1}^m \sum_{j=1}^n   \theta_i^*\lambda_j^*  \left ( \sum_{k=1}^K \sum_{l=1}^L  {1}^{{z}^*}_{ik} {1}^{{w}^*}_{jl} \mu^*_{kl} \right ) \left (\sum_{k'=1}^K \sum_{l'=1}^L q^{{z}}_{ik'} q^{{w}}_{jl'} \log  \mu_{k'l'}  \right ) \\
	& = \sum_{k=1}^K \sum_{l=1}^L \sum_{k'=1}^K \sum_{l'=1}^L \left ( \sum_{i=1}^m \theta_i^*  {1}^{{z}^*}_{ik} q^{{z}}_{ik'} \right ) \left (  \sum_{j=1}^n \lambda_j^* {1}^{{w}^*}_{jl}  q^{{w}}_{jl'} \right ) \mu^*_{kl} \log  \mu_{k'l'}.
	\end{align*}
		\begin{align*}
	\bar{J}_2(q^{{z}},q^{{w}},\mu) & = -\sum_{i=1}^m \sum_{j=1}^n  \theta_i^* \lambda_j^*  \left (\sum_{k'=1}^K \sum_{l'=1}^L q^{{z}}_{ik'} q^{{w}}_{jl'} \mu_{k' l'} \right ) \\
	& =  -\sum_{i=1}^m \sum_{j=1}^n  \theta_i^* \lambda_j^*  \left (\sum_{k'=1}^K \sum_{l'=1}^L q^{{z}}_{ik'} q^{{w}}_{jl'} \mu_{k' l'} \right ) \left (  \sum_{k=1}^K \sum_{l=1}^L  {1}^{{z}^*}_{ik} {1}^{{w}^*}_{jl} \right ) \\
	& = - \sum_{k=1}^K \sum_{l=1}^L \sum_{k'=1}^K \sum_{l'=1}^L \left ( \sum_{i=1}^m \theta_i^*  {1}^{{z}^*}_{ik} q^{{z}}_{ik'} \right ) \left (  \sum_{j=1}^n \lambda_j^* {1}^{{w}^*}_{jl}  q^{{w}}_{jl'} \right )   \mu_{k'l'}.
	\end{align*}
		\begin{align*}
	\bar{J}_1 ({1}^{{z}^*},{1}^{{w}^*},\mu^*) & = \sum_{i=1}^m \sum_{j=1}^n  \theta_i^*\lambda_j^* \left (\sum_{k=1}^K \sum_{l=1}^L {1}^{{z}^*}_{ik}{1}^{{w}^*}_{jl}  \mu^*_{kl}  \log  \mu^*_{kl}  \right )   \\
	& = \sum_{i=1}^m \sum_{j=1}^n  \theta_i^*\lambda_j^* \left (\sum_{k=1}^K \sum_{l=1}^L {1}^{{z}^*}_{ik}{1}^{{w}^*}_{jl}  \mu^*_{kl}  \log  \mu^*_{kl}  \right ) \left (\sum_{k'=1}^K \sum_{l'=1}^L q^{{z}}_{ik'} q^{{w}}_{jl'} \right ) \\
	& = \sum_{k=1}^K \sum_{l=1}^L \sum_{k'=1}^K \sum_{l'=1}^L \left ( \sum_{i=1}^m \theta_i^*  {1}^{{z}^*}_{ik} q^{{z}}_{ik'} \right ) \left (  \sum_{j=1}^n \lambda_j^* {1}^{{w}^*}_{jl}  q^{{w}}_{jl'} \right ) \mu^*_{kl} \log  \mu^*_{kl}.
	\end{align*}
		\begin{align*}
	\bar{J}_2({1}^{{z}^*},{1}^{{w}^*},\mu^*) & = -\sum_{i=1}^m \sum_{j=1}^n  \theta_i^* \lambda_j^*  \left (\sum_{k=1}^K \sum_{l=1}^L  {1}^{{z}^*}_{ik}{1}^{{w}^*}_{jl} \mu^*_{k l} \right ) \\
	& = -\sum_{i=1}^m \sum_{j=1}^n  \theta_i^* \lambda_j^*  \left (\sum_{k=1}^K \sum_{l=1}^L  {1}^{{z}^*}_{ik}{1}^{{w}^*}_{jl} \mu^*_{k l} \right )  \left (\sum_{k'=1}^K \sum_{l'=1}^L q^{{z}}_{ik'} q^{{w}}_{jl'} \right ) \\
	& =-  \sum_{k=1}^K \sum_{l=1}^L \sum_{k'=1}^K \sum_{l'=1}^L \left ( \sum_{i=1}^m \theta_i^*  {1}^{{z}^*}_{ik} q^{{z}}_{ik'} \right ) \left (  \sum_{j=1}^n \lambda_j^* {1}^{{w}^*}_{jl}  q^{{w}}_{jl'} \right )  \mu^*_{kl}.
	\end{align*}
	Therefore,
		\begin{align*}
	& \bar{J}({1}^{{z}^*},{1}^{{w}^*},\mu^*)-\bar{J}(q^{{z}},q^{{w}},\mu) \\
	= \,\, & \sum_{k=1}^K \sum_{l=1}^L \sum_{k'=1}^K \sum_{l'=1}^L \left ( \sum_{i=1}^m \theta_i^*  {1}^{{z}^*}_{ik} q^{{z}}_{ik'} \right ) \left (  \sum_{j=1}^n \lambda_j^* {1}^{{w}^*}_{jl}  q^{{w}}_{jl'} \right ) \textnormal{KL}(\mu^*_{kl},\mu_{k'l'}) \\
	\geq \,\, & r_{mn} \theta_{\textnormal{min}} \lambda_{\textnormal{min}} \sum_{k=1}^K \sum_{l=1}^L \sum_{k'=1}^K \sum_{l'=1}^L \left ( \sum_{i=1}^m   {1}^{{z}^*}_{ik} q^{{z}}_{ik'} \right )  \left (  \sum_{j=1}^n  {1}^{{w}^*}_{jl}  q^{{w}}_{jl'} \right ) \textnormal{KL}(\mu^*_{kl},\mu_{k'l'}) \\
	= \,\, &   mn r_{mn}\theta_{\textnormal{min}} \lambda_{\textnormal{min}}\sum_{k=1}^K \sum_{l=1}^L \sum_{k'=1}^K \sum_{l'=1}^L \mathbb{R}_{ kk'}({1}^{{z}^*},q^{{z}}) \mathbb{R}^{{w}}_{ ll'}({1}^{{w}^*},q^{{w}})  \textnormal{KL}(\mu^*_{kl},\mu_{k'l'}) \geq 0.
	\end{align*}
\end{proof}
We now present a lemma on KL divergence: 
\begin{lemma}\label{lemmaKL}
    $\textnormal{KL}(a,b)=a\log\frac{a}{b}- (a-b)\geq\min\{(a-b)^2/(6b),|a-b|\}$, where $a,\,b>0$.
\end{lemma}
\begin{proof} 
Define $x=\log\frac{a}{b}$. We show below that  $a\log\frac{a}{b}- (a-b)\geq (a-b)^2/(6b) $ when $x\leq 2$, and $a\log\frac{a}{b}- (a-b)\geq |a-b|$ when $x> 2$.

Note that
\begin{align*}
   & a\log\frac{a}{b}- (a-b) \geq (a-b)^2/(6b)\\
   \Longleftrightarrow \quad & \frac{a}{b}\log\frac{a}{b}- (\frac{a}{b}-1)\geq (a-b)^2/(6b^2)\\
   \Longleftrightarrow \quad & e^x x- (e^x-1)\geq \frac16 (e^x-1)^2 \\
   \Longleftrightarrow \quad & 6xe^x-e^{2x}-4e^x+5\geq 0,
\end{align*}
where the last line holds when $x\leq 2$. Moreover, under the condition $x>2$ that implies $a>b$, we have
\begin{align*}
    & a\log\frac{a}{b}- (a-b)\geq |a-b|\\
   \Longleftrightarrow \quad & \frac{a}{b}\log\frac{a}{b}- (\frac{a}{b}-1)\geq \frac{a}{b} -1\\
   \Longleftrightarrow \quad & xe^x\geq 2(e^x-1),
\end{align*}
where the last line trivially holds as $x>2$ and $e^x>e^x-1$.

\end{proof}
We now prove Theorem \ref{thm:separability}. By Lemma \ref{lemmaKL}, we have  
\begin{align*}
\textnormal{KL}(\mu^*_{kl},\mu_{k'l'})\geq \min\{(\mu^*_{kl}-\mu_{k'l'})^2/(6\mu_{\max}),|\mu^*_{kl}-\mu_{k'l'}|\}\geq \min\{(\mu^*_{kl}-\mu_{k'l'})^2/(6\mu_{\max}),6\mu_{\max}\}.
\end{align*}
We define a new metric in $\mathbb{R}^1$ by $|a-b|_{\textnormal{new}}=\min\{|a-b|,6\mu_{\max}\}$. Then we have $\textnormal{KL}(a,b)\geq |a-b|^2_{\textnormal{new}}/(6\mu_{\max})$. For vectors $a$ and $b$, define $\|a-b \|_{\textnormal{new}} = \sqrt{\sum_i |a_i-b_i|^2_{\textnormal{new}} }$. 

We derive the lower bound for $G( q^{{z}},q^{{w}},\mu)$.
Because 
\begin{align*}
\sum_{l'=1}^L \frac{1}{n} \sum_{j=1}^n {1}^{{w}^*}_{jl} q^{{w}}_{jl'} =\frac{1}{n} \sum_{j=1}^n {1}^{{w}^*}_{jl},
\end{align*}
for all $l$, there exists $l'$, denoted by $h(l)$, such that
\begin{align*}
\frac{1}{n} \sum_{j=1}^n {1}^{{w}^*}_{jl} q^{{w}}_{jl'} \geq \frac{1}{L} \frac{1}{n} \sum_{j=1}^n {1}^{{w}^*}_{jl} \geq \frac{1}{L} \tilde{\rho}_{\textnormal{min}}.
\end{align*}
Denote $\tilde{\mu}_{k',l}= \mu_{k',h(l)}$. Then 
\begin{align*}
G( q^{{z}},q^{{w}},\mu) & \geq \frac{1}{L} \hat{\rho}_{\textnormal{min}} \sum_{k=1}^K \sum_{k'=1}^K  \left ( \frac{1}{m} \sum_{i=1}^m  {1}^{{z}^*}_{ik} q^{{z}}_{ik'} \right ) \sum_{l=1}^L \textnormal{KL}(\mu^*_{kl},\tilde{\mu}_{k',l})  \\
& \geq \frac{1}{L} \hat{\rho}_{\min} \sum_{k=1}^K \sum_{k'=1}^K  \left ( \frac{1}{m} \sum_{i=1}^m  {1}^{{z}^*}_{ik} q^{{z}}_{ik'} \right ) 6 \mu_{\textnormal{max}}  \| \mu_{k \cdot}^*- \tilde{\mu}_{k'\cdot } \|_{\textnormal{new}}^2.
\end{align*}
Let 
$d_{\min} =  \min_{k\neq k'}\| \mu_{k \cdot}^*- \mu_{k' \cdot}^*\|_{\textnormal{new}}$. Note that based on $H_3$, $d_{\min}>0$. Furthermore, for any $\tilde{\mu}_{k'\cdot }$ there exists at most one $\mu_{k \cdot}^*$ such that $\|\mu_{k \cdot}^*-\tilde{\mu}_{k' \cdot}\|_{\textnormal{new}}< d_{\min}/2$. 

There are two possible cases:
\begin{itemize}
\item [Case 1:] For each $\mu^*_{k \cdot}$, there exists one and only one $\tilde{\mu}_{k' \cdot }$ such that $\|\mu_{k \cdot }^*-\tilde{\mu}_{k' \cdot}\|_{\textnormal{new}}< d_{\min}/2$. 
\item [Case 2:] There exists some $\mu^*_{k \cdot }$ such that no $\tilde{\mu}_{k' \cdot}$ is within its $d_{\min}/2$-radius.
\end{itemize}
The one-to-one correspondence in Case 1 induces a permutation $s$ on $\{1,...,K\}$.  Case 1 implies $\|\mu_{k \cdot}^*-\tilde{\mu}_{k' \cdot}\|_{\textnormal{new}}^2 \geq d_{\min}^2/4$ for $k \neq s(k')$.
\begin{align*}
 & \sum_{k=1}^K \sum_{k'=1}^K  \left ( \frac{1}{m} \sum_{i=1}^m  {1}^{{z}^*}_{ik} q^{{z}}_{ik'} \right ) \| \mu_{k \cdot}^*- \tilde{\mu}_{k'\cdot } \|_{\textnormal{new}}^2 \\
\geq \,\, & \sum_{k'=1}^K \sum_{k \neq s(k')} \left ( \frac{1}{m} \sum_{i=1}^m  {1}^{{z}^*}_{ik} q^{{z}}_{ik'} \right ) \| \mu_{k \cdot}^*- \tilde{\mu}_{k'\cdot } \|_{\textnormal{new}}^2 \\
\geq \,\, & \frac{d_{\min}^2}{4} \sum_{k'=1}^K \sum_{k \neq s(k')} \left ( \frac{1}{m} \sum_{i=1}^m  {1}^{{z}^*}_{ik} q^{{z}}_{ik'} \right ) \geq  \frac{d_{\min}^2}{4} M_\textnormal{row}(q^{{z}}).
\end{align*}
In Case 2, let $k$ be the class label such that $\|\mu_{k \cdot}^*-\tilde{\mu}_{k' \cdot }\|_{\textnormal{new}}^2\geq d_{\min}^2/4$ for all $k'$. 
\begin{align*}
 & \sum_{k=1}^K \sum_{k'=1}^K  \left ( \frac{1}{m} \sum_{i=1}^m  {1}^{{z}^*}_{ik} q^{{z}}_{ik'} \right ) \| \mu_{k \cdot}^*- \tilde{\mu}_{k'\cdot } \|_{\textnormal{new}}^2 \\
 \geq \,\, &  \sum_{k'=1}^K  \left ( \frac{1}{m} \sum_{i=1}^m  {1}^{{z}^*}_{ik} q^{{z}}_{ik'} \right ) \| \mu_{k \cdot}^*- \tilde{\mu}_{k'\cdot } \|_{\textnormal{new}}^2 \\
 \geq \,\, & \frac{d_{\min}^2}{4}  \frac{1}{m} \sum_{i=1}^m  {1}^{{z}^*}_{ik} \geq   \frac{d_{\min}^2}{4} \tilde{\rho}_{\min} M_\textnormal{row}(q^{{z}}).
\end{align*}

In summary, $G( q^{{z}},q^{{w}},\mu) \geq  C_1  M_\textnormal{row}(q^{{z}})$. Based on the same argument, $G( q^{{z}},q^{{w}},\mu) \geq  C_2  M_\textnormal{col}(q^{{w}})$.

\subsection{Proof of Theorem \ref{thm:consistency}}
Let $\Phi^*=({\pi}^*,{\rho}^*,\mu^*)$. By Theorems \ref{thm:uniform} and \ref{thm:separability}, if $(mn r_{mn} \epsilon^2)/(m+n) \rightarrow \infty$,  
\begin{align*}
 & \,\, \mathbb{P} \left (M_\textnormal{row}(\hat{q}^{{z}})\geq \epsilon \big \vert z^*, w^* \right ) \\
 \leq  & \,\, \mathbb{P} \left ( \bar{J}({1}^{{z}^*},{1}^{{w}^*},\mu^*)-\bar{J}(\hat{q}^{{z}},\hat{q}^{{w}}, \hat{\mu}) \geq  C_1 mn r_{mn}  \epsilon  \big \vert z^*,w^*   \right ) \\
 = & \,\, \mathbb{P}\left (\bar{J}({1}^{{z}^*},{1}^{{w}^*},\mu^*)-\hat{J}({1}^{{z}^*},{1}^{{w}^*},\Phi^*)+\hat{J}({1}^{{z}^*},{1}^{{w}^*},\Phi^*)-\hat{J}(\hat{q}^{{z}},\hat{q}^{{w}},\hat{\Phi}) \right.  \\
	& \,\,  \left. +\hat{J}(\hat{q}^{{z}},\hat{q}^{{w}},\hat{\Phi})-\bar{J}(\hat{q}^{{z}},\hat{q}^{{w}},\hat{\mu})  \geq   C_1 mn r_{mn}  \epsilon \big \vert  {z}^*,{w}^*  \right ) \\
 \leq &  \,\, \mathbb{P}\left (\bar{J}({1}^{{z}^*},{1}^{{w}^*},\mu^*)-\hat{J}({1}^{{z}^*},{1}^{{w}^*},\Phi^*) +\hat{J}(\hat{q}^{{z}},\hat{q}^{{w}},\hat{\Phi})-\bar{J}(\hat{q}^{{z}},\hat{q}^{{w}},\hat{\mu}) \geq  C_1 mn r_{mn} \epsilon   \big \vert  {z}^*,{w}^*\right ) \\
 \leq & \,\, \mathbb{P}\left (|\bar{J}({1}^{{z}^*},{1}^{{w}^*},\mu^*)-\hat{J}({1}^{{z}^*},{1}^{{w}^*},\Phi^*)|  \geq  (C_1/2) mn r_{mn} \epsilon   \big \vert  {z}^*,{w}^*\right )\\
     & \,\, + \mathbb{P}\left (|\hat{J}(\hat{q}^{{z}},\hat{q}^{{w}},\hat{\Phi})-\bar{J}(\hat{q}^{{z}},\hat{q}^{{w}},\hat{\mu})|  \geq  (C_1/2) mn r_{mn} \epsilon   \big \vert  {z}^*,{w}^*\right ) \rightarrow 0.
\end{align*}
It implies, for $\epsilon =\left ( \frac{mnr_{mn}}{m+n}\right)^{-1/2+\delta} \epsilon_1$, where $\delta$ and $\epsilon_1$ are positive constants, 
\begin{align*}
\mathbb{P} \left (M_\textnormal{row}(\hat{q}^{{z}})\geq \epsilon \big \vert z^*, w^* \right ) \rightarrow 0; 
\end{align*}
in other words, 
\begin{align*}
M_\textnormal{row}(\hat{q}^{{z}})=o_p \left ( \left ( \frac{mnr_{mn}}{m+n}\right)^{-1/2+\delta} \right).
\end{align*}	
Similarly, 
\begin{align*}
 M_\textnormal{col}(\hat{q}^{{w}})=o_p \left ( \left ( \frac{mnr_{mn}}{m+n}\right)^{-1/2+\delta} \right).
\end{align*}

\subsection{Proof of Theorem \ref{thm:consistency_ber}}
We only need to prove the uniform convergence of $\hat{J}(q^{{z}} ,q^{{w}},\Phi)$ to $\bar{J}(q^{{z}},q^{{w}},\mu)$ under the Bernoulli model, which relies on the following lemma:
\begin{lemma}\label{thm:general_ber}
	Let $\{X_{ij}\}$ be independent Bernoulli variables with mean $E[X_{ij}] \leq r_{mn} C$. Then for all $\epsilon>0$,
	\begin{align*}
	\mathbb{P} \left ( \max_{0\leq u_i \leq 1, i=1,...,m,  0 \leq v_j \leq 1,j=1,...,n}  \left |   \sum_{i=1}^m \sum_{j=1}^n  (X_{ij}-E[X_{ij}])  u_i v_j  \right | \geq mn r_{mn}  \epsilon \right )  \\
 \leq  2^{m+n+1}  \exp \left ( -\frac{m n r_{mn} \epsilon^2}{4 \max(C,\epsilon/3)} \right ).
	\end{align*}
\end{lemma}
\begin{proof}
According to the Bernstein inequality, for $u_i\in \{0, 1\},i=1,...,m, v_j \in \{0,1\},j=1,...,n$,
\begin{align*}
& \mathbb{P} \left (  \left |   \sum_{i=1}^m \sum_{j=1}^n  (X_{ij}-E[X_{ij}])  u_i v_j  \right | \geq mn r_{mn} \epsilon \right ) \\
\leq & \,\, 2 \exp \left ( -\frac{m^2 n^2 r_{mn}^2 \epsilon^2}{2(\sum_{ij}  \textnormal{Var}[X_{ij}]+mn r_{mn} \epsilon/3)} \right ) \leq 2 \exp \left ( -\frac{m n r_{mn} \epsilon^2}{4\max(C,\epsilon/3)} \right ).
\end{align*}
The rest of the proof is identical to that of Lemma \ref{thm:general_poisson}.

\end{proof}

\section{Application to SMS spam data set}\label{sec:spam}

We apply the proposed method to an SMS spam data set collected by \cite{almeida2011contributions}. The data set has a total of 4827 SMS legitimate messages (labeled as ``ham'') and a total of a total of 747 spam messages (labeled as ``spam''). We followed a standard procotol for data preprocessing (https://kharshit.github.io/blog/2017/08/25/email-spam-filtering-text-analysis-in-r) using the R package \textsf{tm}: we removed punctuation and stopwords (such as ``that'') from the messages, and only kept words appearing in at least 1\% of the messages. This results in a total of 4938 messages, with 4211 labeled as ``ham'' and 727 labeled as ``spam'', and a vocabulary size of 139. The messages were then encoded in a $4938 \times 139$ matrix $A=[A_{ij}]$ where each entry $A_{ij}$ represents the count of the $j$-th word in the $i$-th message. The data matrix, therefore, represents a bipartite network with integer weights.

The data set typically serves as a benchmark for supervised learning in spam filtering. We took an unsupervised learning approach; that is, we applied biclustering methods to this data set and compared the estimated labels on the messages with the true labels. We applied the three aforementioned methods---SC, PL, and DC-LBM---to the data set with $K=2$ (since the messages are classified as ``ham'' or ``spam'') and $L=2, 3, 4, 5, 6$. We reported two metrics---ARI and accuracy---for evaluating the performances. The latter is the fraction of estimated labels that match the true labels, allowing for a permutation of the two classes.  The results for SC, PL, and DC-LBM are reported in Table \ref{table:spam}. Note that the performance of SC on row clustering does not depend on $L$.  The best performance of PL is achieved at $L=3$, and the best performance of  DC-LBM is achieve at $L=5$. For all values of $L$, DC-LBM consistently outperforms PL and SC in terms of both ARI and accuracy. This pattern aligns with the findings in the simulation studies. 
\begin{table}[ht!]
\centering
\begin{tabular}{|l|c|c|c|c|c|}
\hline
$L$ & 2 & 3 & 4 & 5 & 6 \\
\hline
ARI (SC) & \multicolumn{5}{c|}{0.419}  \\
\hline
ARI (PL)  &  0.085 & 0.165 & 0.112 & 0.125 & 0.101\\
\hline
ARI (DC-LBM)  &  0.634 & 0.617 & 0.719 & 0.729  & 0.644 \\
\hline 
Accuracy (SC) & \multicolumn{5}{c|}{0.851}  \\
\hline
Accuracy (PL)  & 0.654 & 0.716 & 0.672 & 0.683 & 0.661 \\
\hline 
Accuracy (DC-LBM) & 0.923 & 0.916 & 0.944 & 0.946 & 0.924 \\
\hline 
\end{tabular}
\caption{ARI and accuracy of SC, PL. and DC-LBM on the SMS spam data set.} \label{table:spam}
\end{table}

 \newcommand{\noop}[1]{}

\end{document}